\documentclass[11pt]{article}

\usepackage[final]{acl}

\usepackage{times}
\usepackage{latexsym}

\usepackage[T1]{fontenc}

\usepackage[utf8]{inputenc}

\usepackage{microtype}

\usepackage{inconsolata}

\usepackage{graphicx}

\usepackage{amssymb}
\usepackage{multirow}
\usepackage{makecell}
\usepackage{amsmath}
\usepackage{booktabs}
\usepackage{diagbox}
\usepackage{pifont}
\usepackage[most]{tcolorbox}

\title{M$^3$-VQA: A Benchmark for Multimodal, Multi-Entity, Multi-Hop \\ Visual Question Answering}

\author{
 \textbf{Jiatong Ma\textsuperscript{1,2}\thanks{Equal Contribution.}},
 \textbf{Longteng Guo\textsuperscript{1}\footnotemark[1]},
 \textbf{Yuchen Liu\textsuperscript{1,2}},
 \textbf{Zijia Zhao\textsuperscript{1,2}},
\\
 \textbf{Dongze Hao\textsuperscript{1,2}},
 \textbf{Xuanxu Lin\textsuperscript{1,2}},
 \textbf{Jing Liu\textsuperscript{1,2}\thanks{Corresponding author.}}
\\
\\
 \textsuperscript{1}
Institute of Automation, Chinese Academy of Sciences,
\\
 \textsuperscript{2}School of Artificial Intelligence, University of Chinese Academy of Sciences, 
\\
   \texttt{majiatong2025@ia.ac.cn, \{longteng.guo,jliu\}@nlpr.ia.ac.cn}
}

\begin{document}
\maketitle
\begin{abstract}
We present M$^3$-VQA, a novel knowledge-based Visual Question Answering (VQA) benchmark, to enhance the evaluation of multimodal large language models (MLLMs) in fine-grained multimodal entity understanding and complex multi-hop reasoning. Unlike existing VQA datasets that focus on coarse-grained categories and simple reasoning over single entities, M$^3$-VQA introduces diverse multi-entity questions involving multiple distinct entities from both visual and textual sources. It requires models to perform both sequential and parallel multi-hop reasoning across multiple documents, supported by traceable, detailed evidence and a curated multimodal knowledge base. We evaluate 16 leading MLLMs   under three settings: without external knowledge, with gold evidence, and with retrieval-augmented input. The poor results reveal significant challenges for MLLMs in knowledge acquisition and reasoning. Models perform poorly without external information but improve markedly when provided with precise evidence. Furthermore, reasoning-aware agentic retrieval surpasses heuristic methods, highlighting the importance of structured reasoning for complex multimodal understanding. M$^3$-VQA presents a more challenging evaluation for advancing the multimodal reasoning capabilities of MLLMs. Our code and dataset are available at \url{https://github.com/CASIA-IVA-Lab/M3VQA}.
\end{abstract}

\section{Introduction}
With the rapid advancement of Multimodal Large Language Models (MLLMs), Visual Question Answering (VQA) has emerged as a widely used benchmark for evaluating open-ended multimodal understanding. These models have demonstrated impressive progress in surface-level perception and commonsense reasoning. However, real-world applications often require more than basic comprehension—they demand the ability to perform complex reasoning involving multiple, fine-grained visual entities. For example, a question might require identifying specific people, brands, or animal species within an image and then integrating their attribute and relationship information through multi-entity, multi-hop reasoning to arrive at the correct answer.

\begin{figure*}
  \centering
  \includegraphics[width=0.9\textwidth]{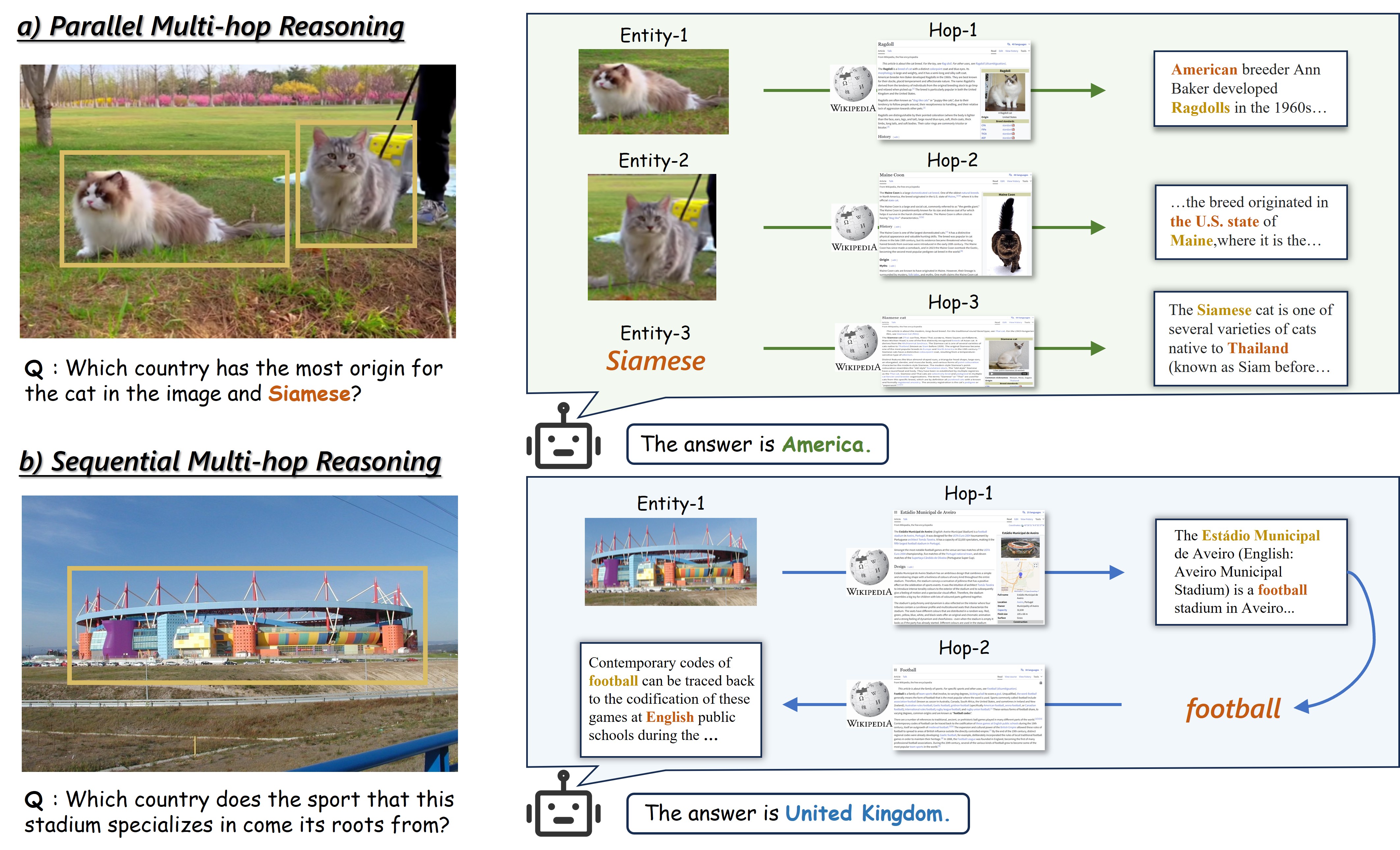}
  \caption{Examples of parallel and sequential multi-hop reasoning in M$^3$-VQA.}
  \label{comparison}
\vspace{-10pt}
\end{figure*}

Existing knowledge-based VQA benchmarks, such as OK-VQA \cite{marino2019ok} and A-OKVQA \cite{schwenk2022okvqa}, have advanced the field by introducing questions that require external knowledge beyond the image content. Nonetheless, these datasets primarily focus on relatively coarse-grained, basic-level categories and entities, and typically emphasize reasoning over single entities. While recent datasets like S3VQA \cite{jain2021select}, ViQuAE  \cite{lerner2022viquae}, KVQA \cite{shah2019kvqa}, EVQA \cite{mensink2023encyclopedic}, InfoSeek \cite{chen2023can} and Dyn-VQA \cite{li2024benchmarking} have attempted to incorporate finer-grained factual knowledge, they mostly address relatively simple scenarios. The questions are often answerable with only one or two reasoning steps and involve a single primary entity, limiting their ability to fully challenge and evaluate the complex multimodal reasoning skills of modern MLLMs.

To address these limitations, we introduce M$^3$-VQA, a novel and challenging benchmark designed to significantly advance the evaluation of knowledge-based Visual Question Answering. M$^3$-VQA specifically targets the model's capability in fine-grained, multimodal entity understanding and sophisticated multi-hop reasoning. Our benchmark introduces three key features:

\textbf{(1) Diverse Multi-Entity Questions:} M$^3$-VQA incorporates a variety of fine-grained named entities—including architectural landmarks, individual person names, brands, and animal species—sourced from both visual scenes and textual information. Each question in the dataset involves multiple distinct entities, closely reflecting realistic multimodal interactions.

\textbf{(2) Complex Multi-Hop Reasoning:} Questions in M$^3$-VQA require models to retrieve and integrate information from multiple documents or external knowledge sources, promoting extensive cross-modal and cross-document reasoning rather than reliance on single-step inference. Crucially, the dataset encompasses both parallel reasoning tasks—where multiple entities are independently analyzed before synthesizing an answer—and sequential reasoning tasks—where entities form a chain that models must sequentially traverse to reach the solution.

\textbf{(3) Traceable Supporting Evidence:} Each reasoning step in M$^3$-VQA is explicitly supported by detailed, grounded evidence, allowing transparent traceability of model reasoning processes. Additionally, we provide a carefully curated multimodal knowledge base to support retrieval-augmented evaluation protocol.

This comprehensive design enables rigorous evaluation of MLLMs, both in standalone settings and scenarios enhanced by retrieval mechanisms, effectively pushing the boundaries of real-world multimodal understanding and reasoning.

We evaluate 16 leading MLLMs under three settings: without evidence, with gold evidence, and with retrieval from an external knowledge base. Our experiments reveal the following key findings:
\begin{itemize}

\item \textbf{MLLMs Perform Poorly without External Knowledge:} When restricted to only the image and question, models consistently underperform, with a maximum accuracy of 32.6\%. This reveals a fundamental limitation of current MLLMs in acquiring and applying background knowledge solely from their internal representations.

\item \textbf{Precise Evidence Significantly Boosts Reasoning Accuracy:} When gold supporting evidence is provided, model performance improves substantially. This indicates that even advanced MLLMs remain heavily reliant on well-structured external information to support complex multi-entity reasoning.

\item \textbf{Reasoning-Aware Agentic Retrieval Outperforms Heuristic Approach:} While retrieval-augmented approaches boost performance, we find that agentic retrieval—featuring explicit reasoning and iterative planning—outperforms heuristic retrieval. This underscores the importance of structured reasoning strategies in tackling multi-entity, multi-hop questions.

\end{itemize}

\begin{table*}
  
  \centering
\resizebox{\linewidth}{!}{
  \small
  \begin{tabular}{lcccccccccccc}
\toprule
\textbf{}

& \textbf{OKVQA}
& \textbf{S3VQA}
& \textbf{ViQuAE} 
& \textbf{KVQA} 
& \textbf{InfoSeek} 
& \textbf{EVQA} 
& \textbf{Dyn-VQA} 
& \textbf{M$^3$-VQA } \\

\midrule


Multi-Entity & $\times$ & $\times$ & $\times$ & $\checkmark$$\checkmark$ & $\times$ & $\times$ & $\checkmark$ & $\checkmark$$\checkmark$ \\

Fine-Grained Entity & $\times$ & $\checkmark$ & $\checkmark$$\checkmark$ & $\checkmark$$\checkmark$ & $\checkmark$$\checkmark$ & $\checkmark$$\checkmark$ & $\checkmark$$\checkmark$ & $\checkmark$$\checkmark$ \\

Entity Diversity & $\checkmark$$\checkmark$ & $\checkmark$$\checkmark$ & $\checkmark$$\checkmark$ & $\times$ & $\checkmark$$\checkmark$ & $\checkmark$ & $\checkmark$$\checkmark$ & $\checkmark$$\checkmark$ \\


Multi-Hop & $\times$ & $\times$ & $\times$ & $\checkmark$$\checkmark$ & $\times$ & $\checkmark$ & $\checkmark$ & $\checkmark$$\checkmark$ \\

Reasoning Mode\\

~~~~$\hookrightarrow$Sequential & $\times$ & $\times$ & $\times$ & $\checkmark$ & $\times$ & $\checkmark$ & $\checkmark$ & $\checkmark$$\checkmark$ \\

~~~~$\hookrightarrow$Parallel & $\times$ & $\times$ & $\times$ & $\checkmark$$\checkmark$ & $\times$ & $\times$ & $\checkmark$ & $\checkmark$$\checkmark$ \\

Controlled Knowledge Base\\
~~~~$\hookrightarrow$Traceable Answer & $\times$ & $\times$ & $\checkmark$ & $\times$ & $\times$ & $\checkmark$ & $\times$ & $\checkmark$$\checkmark$ \\

~~~~$\hookrightarrow$Free-Form KB & $\times$ & $\times$ & $\checkmark$$\checkmark$ & $\times$ & $\checkmark$$\checkmark$ & $\checkmark$$\checkmark$ & $\times$ & $\checkmark$$\checkmark$ \\


\midrule
Answer Type & Open & Open & Open & Open & Open & Open & Open & Open \\
\#\{I, Q, A\} & 14K & 7K & 3K & 183K & 1M & 1M & 1K & 13K \\


\bottomrule
\end{tabular}
}

  \caption{Comparison between our M$^3$-VQA and previous VQA datasets.}
\vspace{-10pt}
\label{datasets_comparison}
\end{table*}

\section{Related Work}

Real-world VQA often requires external knowledge, leading to growing interest in retrieval for VQA. Early works like KB-VQA \cite{wang2015explicit}, FVQA \cite{wang2017fvqa}, and KVQA \cite{shah2019kvqa} relied on structured knowledge bases, which limited the scope of information. In contrast, we aim to build a knowledge base from free-form text and images (e.g., Wikipedia). KVQA also focuses solely on facial recognition, lacking broad entity coverage. Datasets such as OKVQA \cite{marino2019ok} and A-OKVQA \cite{schwenk2022okvqa} primarily require commonsense knowledge and rarely involve detailed attributes of fine-grained entities, thus often eliminating the need for retrieval. While S3VQA \cite{jain2021select} and ViQuAE \cite{lerner2022viquae} address fine-grained attributes, S3VQA lacks a knowledge base, and ViQuAE is small (3.6k samples). Recent datasets like EVQA \cite{mensink2023encyclopedic} and Infoseek \cite{chen2023can} do include fine-grained entity attributes and scale to millions of examples, solving the size issue. However, they still focus on single-entity images and questions that can be answered within two retrieval hops, failing to capture the complexity of multi-entity interaction and multi-hop reasoning in real-world scenarios. Dyn-VQA \cite{li2024benchmarking} recently introduced multi-hop questions but is very small (only 387 samples) and lacks a knowledge base. Our proposed M$^3$-VQA advances prior work in multiple dimensions:
(1) It uniquely integrates multimodal inputs, multi-entity interactions, and multi-hop reasoning, offering a more challenging comprehensive evaluation environment.
(2) It includes a dedicated knowledge base and gold evidence annotations to support each answer, enabling precise assessment of reasoning capabilities.
See Appendix~\ref{appx:A} for more.

\section{M$^3$-VQA Task Definition}


The M$^3$-VQA task can be formulated as a function \( f: (I, Q, K) \rightarrow A \), where \( I \) denotes an input image, \( Q \) represents a question, \( K \) is an external Wikipedia knowledge base, and \( A \) is the answer. Specifically, each image \( I \) may contain multiple visual entities. Each question \( Q \) is expressed in text and may comprise multiple textual entities or sub-questions. The external Wikipedia knowledge base \( K \) includes multimodal pages with both text and images. The answer \( A \) can take various forms, including strings (words or phrases) and temporal values (e.g., years or dates). For each sample, there exists a reference evidence chain  \( E \) that serves as a guide for the reasoning process. We assess the difficulty of the task from two key dimensions: 
\textbf{(1) Entity Complexity:} defined as the total number of visual and textual entities involved in \( Q \) and \( I \). 
\textbf{(2) Hop Complexity:} measured by the number of nodes in the reference evidence chain \( E \) required to answer the question \( Q \).

Specifically, we divided M$^3$-VQA task into 2 sub-task based on the reasoning pattern of multi-hop  evidence chain: \textbf{(1) Parallel Multi-hop Reasoning} refers to a multi-hop inference process conducted in a parallel manner, where multiple entities within the image and question are reasoned about simultaneously. The final answer is derived from the integrated understanding of the individual results obtained from reasoning over each entity in parallel. For example, in the first image of Figure~\ref{comparison}, we can independently retrieve information about the two cats in the image and the Siamese mentioned in the question to determine their countries of origin, and then identify which country appears most frequently. \textbf{(2) Sequential Multi-hop Reasoning} refers to a multi-hop inference process structured in a serial manner, where the input of each subsequent hop is derived from the output of the previous hop. The final result is obtained from the end of this sequential chain. For instance, in the second image of Figure~\ref{comparison}, one must first retrieve information about the stadium to determine which sport it is used for and, subsequently, retrieve information about that sport before determining its country of origin.

\section{M$^3$-VQA Dataset}

\subsection{Dataset Construction}
\noindent\textbf{Image and Entity Annotation.}
Our dataset construction begins with collecting images containing multiple fine-grained entities, leveraging existing image datasets that already provide detailed entity annotations. Specifically, we utilize fine-grained image datasets such as CelebTo \cite{zhong2018compact}, FGVD \cite{khoba2022fine}, FlickrLogos-47 \cite{romberg2011scalable}, LogosInTheWild \cite{tuzko2017open}, Menu-match \cite{beijbom2015menu}, Oktoberfest \cite{ziller2019oktoberfest}, UEC-FoodPix \cite{okamoto2021uec} and UNIMIB2016 \cite{ciocca2016food}, which include precise labels for person, vehicle, logo, and food within images. These original entity annotations form the foundation for our benchmark, ensuring reliable multi-entity grounding. To further increase dataset diversity, we also incorporate the single-entity, single-hop images from knowledge-based VQA datasets such as EVQA~\cite{mensink2023encyclopedic} and InfoSeek~\cite{chen2023can}, and systematically extend their questions to multi-hop settings. 
Additional images were manually collected by human annotators via web search engines to supplement fine-grained multi-entity scenes.

\noindent\textbf{Entity Linking and Expansion.}
Starting from these images and their entity annotations, we map each entity label to a corresponding Wikidata identifier~\cite{vrandevcic2014wikidata} through automatic string matching. 
For ambiguous or uncertain mappings, human annotators performed manual verification based on image-label pairs to ensure accurate entity linking. Leveraging Wikidata IDs, we retrieved corresponding Wikipedia articles, enabling multimodal grounding between images, structured Wikidata knowledge, and multimodal Wikipedia pages. This linkage facilitates transparent reasoning and retrieval of rich external knowledge for each entity.

\noindent\textbf{Parallel Multi-Hop Question Generation.}
To generate questions requiring parallel multi-hop reasoning, we exploited Wikidata’s structured knowledge triples (subject, relation, object). Human annotators designed natural language question templates for diverse relation types, employing placeholders for high-level visual categories (e.g., “person”, “brand”). We used SPARQL queries to extract answers from Wikidata which were processed through the question templates to form final answers.
Images were then paired with these question-answer pairs to construct IQA triplets. Subsequently, additional textual entities of the same category were incorporated into questions to increase complexity, with corresponding sub-answers retrieved and aggregated into the final answers. This process resulted in questions that require models to independently reason about multiple entities before synthesizing a response, supporting robust evaluation of parallel multi-hop capabilities. To increase linguistic variety, some template-based questions were paraphrased with LLMs.

\noindent\textbf{Sequential Multi-Hop Question Generation.}
For sequential multi-hop reasoning questions, we adopted the concept of “bridging entities”~\cite{yang2018hotpotqa}, where the answer to one sub-question serves as the basis for a subsequent question. We automatically identified bridging entities within answers, filtering out non-informative terms. Leveraging large language models (LLMs), we generated candidate follow-up questions linked to these bridging entities. These candidates were validated by ensuring their answers matched the original grounding information, with invalid or ambiguous questions discarded. Complex multi-hop questions were constructed by chaining multiple sequential reasoning steps.

\noindent\textbf{Knowledge Base and Evidence Annotation.}
We constructed a multimodal knowledge base by extracting Wikipedia pages—including both textual content and images—from a snapshot comprising approximately 2 million entities, to facilitate retrieval-augmented evaluation. Each sub-question is linked to its corresponding Wikipedia article, where the exact sentence containing the answer was identified and annotated as gold evidence. Additionally, the section indices of supporting sentences were recorded to enable precise traceability of the reasoning process.

\noindent\textbf{Quality Control.}
To maintain high data quality, we employed a strict annotation protocol involving trained human annotators who underwent screening before participating. Quality control measures included multiple rounds of filtering to exclude low-quality images, irrelevant or overly generic entities, and questions with insufficient length or ambiguous answers. We ensured a balanced distribution across question types and reasoning complexities through strategic sub-sampling. The final M$^3$-VQA dataset offers a diverse and challenging benchmark tailored to advanced multimodal, multi-entity, multi-hop reasoning in knowledge-based VQA.  See Appendix~\ref{appx:D} for more.

\subsection{Dataset Statistics}

As shown in Tables~\ref{dataset_statistics}, our dataset contains 13k (I, Q, A) triples, 10,565 unique images, and 7,611 unique questions. The ratio of unique images is 80\%, and the ratio of unique questions is 58\%, demonstrating the diversity of the dataset. 
In terms of hop count, the dataset includes 1,092 single-hop questions, 3,181 two-hop questions, 4,546 three-hop questions, and 4,306 questions with four or more hops. In terms of the number of entities, the dataset includes 2,992 single-entity questions, 2,183 two-entity questions, 3,644 three-entity questions, and 4,306 questions with four or more entities. Table~\ref{datasets_comparison} also presents a comparative analysis between M$^3$-VQA and other knowledge-based VQA datasets. The dataset includes numerous visual entities covering a wide range of categories such as person, animal, plant, vehicle, food, logo, building, and landmark.
See Appendix~\ref{appx:E} for more.

\begin{table}
\centering
\small
\begin{tabular}{lcr}
      \toprule
      \multicolumn{2}{l}{\textbf{Statistic}} & \textbf{Number} \\
      \midrule
      \multicolumn{2}{l}{(I, Q, A) triplets} & 13125 \\
      \multicolumn{2}{l}{Unique I} & 10565 (80.5\%) \\
      \multicolumn{2}{l}{Unique Q} & 7611 (58.0\%) \\
      \multicolumn{2}{l}{Unique entity} & 4645 \\
      \multicolumn{2}{l}{Average length of Q} & 14.2 \\
      \multicolumn{2}{l}{Average \#hop} & 3.1 \\
      \multicolumn{2}{l}{Average \#entity} & 2.8 \\
      \multicolumn{2}{l}{Average \#Evidence} & 3.1 \\
      \midrule
      \multirow{4}{*}{\#Hop}
      & 1  & 1092 (8.3\%) \\
      & 2  & 3181 (24.2\%) \\
      & 3  & 4546 (34.6\%) \\
      & 4+ & 4306 (32.8\%) \\
      \midrule
      \multirow{4}{*}{\#Entity}
      & 1  & 2992 (22.8\%) \\
      & 2  & 2183 (16.6\%) \\
      & 3  & 3644 (27.8\%) \\
      & 4+ & 4306 (32.8\%) \\
      \bottomrule
    \end{tabular}
\caption{Statistics of M$^3$-VQA.}
\label{dataset_statistics}
\end{table}


\subsection{Evaluation Metrics}
Given the prevalence of multi-answer questions in the dataset, we follow EVQA \cite{mensink2023encyclopedic} to use the Intersection over Union (IoU) between the predicted and ground-truth answer sets as the accuracy for each question. The final score is the arithmetic mean across all questions. Following previous work \cite{goyal2017making, marino2019ok, chen2023can}, we apply exact match using pre-collected answer variations for string-based answers, and we use a relaxed match allowing a one-year margin for time-based questions, considering the estimation nature of historical dates.

\section{Experiments}

\subsection{Evaluation Settings}

\begin{table*}[t]
  \centering
  \vspace{-7pt}
  \small
  \begin{tabular}{lccccccccc}
\toprule
 \multirow{2}{*}{\textbf{Model}} 
& \multicolumn{4}{c}{\textbf{Hop}} 
& \multicolumn{4}{c}{\textbf{Entity}} 
& \multirow{2}{*}{\textbf{All}}  \\
\cmidrule(lr){2-5}
\cmidrule(lr){6-9}
& \textbf{1} & \textbf{2} & \textbf{3} & \textbf{4+} 
  & \textbf{1} & \textbf{2} & \textbf{3} & \textbf{4+} 
  &  \\
\midrule
Qwen2.5-VL-3B-Instruct \cite{bai2025qwen2}        & 37.0 & 17.4 & 17.6 & 16.0 & 24.5 & 18.6 & 17.0 & 16.0 & 18.7  \\
Qwen2.5-VL-7B-Instruct \cite{bai2025qwen2}        & 34.9 & 20.7 & 21.6 & 21.7 & 24.1 & 24.3 & 21.1 & 21.7 & 22.5  \\
Qwen2.5-VL-32B-Instruct \cite{bai2025qwen2}       & 46.8 & 24.9 & 29.8 & 25.6 & 30.1 & 29.4 & 30.7 & 25.6 & 28.7  \\
Qwen2.5-VL-72B-Instruct \cite{bai2025qwen2}       & 47.9 & \textbf{29.5} & \textbf{34.2} & \textbf{29.2} & 30.8 & \textbf{35.5} & \textbf{36.2} & \textbf{29.2} & \textbf{32.6}  \\
Qwen2-VL-7B-Instruct \cite{wang2024qwen2}          & 28.7 & 17.6 & 20.5 & 24.3 & 20.5 & 20.9 & 20.1 & 24.3 & 21.7  \\
InternVL2.5-4B \cite{chen2024expanding}              & 35.0 & 13.3 & 18.0 & 19.7 & 18.4 & 17.0 & 19.2 & 19.7 & 18.8  \\
InternVL2.5-8B \cite{chen2024expanding}               & 26.1 & 16.9 & 19.4 & 17.4 & 18.5 & 19.6 & 19.9 & 17.4 & 18.7  \\
InternVL2.5-26B \cite{chen2024expanding}              & 27.4 & 13.8 & 15.2 & 17.0 & 17.9 & 15.4 & 15.3 & 17.0 & 16.5  \\
InternVL2.5-38B \cite{chen2024expanding}              & 42.2 & 24.4 & 30.6 & 26.0 & 28.7 & 28.9 & 31.2 & 26.0 & 28.6  \\
InternVL2.5-78B	\cite{chen2024expanding}             & 43.5 & 27.2 & 33.4 & 29.1 & 29.8 & 32.5 & 34.4 & 29.1 & 31.3   \\
LLaVA-OneVision-7B \cite{li2024llava} & 35.0 & 17.5 & 21.6 & 22.5 & 21.3 & 20.9 & 22.7 & 22.5 & 22.0  \\
DeepSeek-VL2 (4.5B) \cite{wu2024deepseek}               & 40.2 & 20.3 & 24.9 & 22.7 & 24.6 & 24.0 & 26.2 & 22.7 & 24.3  \\
MiniCPM-V-2.6 (8B) \cite{yao2024minicpm}                & 31.5 & 19.1 & 23.1 & 25.1 & 19.2 & 23.1 & 25.3 & 25.1 & 23.5  \\
GPT-4o $^{*}$ \cite{hurst2024gpt} & \textbf{51.6} & 23.3 & 24.3 & 27.9 & \textbf{34.4} & 25.9 & 22.3 & 27.9 & 27.5 \\

\midrule
\multicolumn{4}{l}{\textit{Question and Image Description as Input}} \\
Llama-3.1-8B-Instruct (\textit{text}) \cite{grattafiori2024llama}        & 26.2 & 15.9 & 21.7 & 21.5 & 18.6 & 19.3 & 21.9 & 21.5 & 20.6  \\
Qwen2.5-7B-Instruct (\textit{text}) \cite{yang2024qwen2}           & 24.0 & 12.9 & 18.7 & 17.4 & 16.7 & 16.1 & 18.4 & 17.4 & 17.3  \\

\bottomrule
\end{tabular}
\vspace{-5pt}
  \caption{Performance comparison of various models under Original setting. * indicates that the model may have a lower accuracy due to its tendency to refuse to answer human queries. See Appendix~\ref{appx:H} for more model results.}
  \label{model_performance_original}
\vspace{-10pt}
\end{table*}

We benchmark a wide range of models, including advanced closed-source model GPT-4o \cite{hurst2024gpt}, and open-source models such as Qwen2.5-VL-72B-Instruct \cite{bai2025qwen2}. To compare visual models from different providers, we evaluate 7B/8B versions of Qwen2.5-VL, Qwen2-VL \cite{wang2024qwen2}, InternVL2.5 \cite{chen2024expanding}, Llava-OneVision \cite{li2024llava}, DeepSeek-VL2 \cite{wu2024deepseek} and MiniCPM-V-2.6 \cite{yao2024minicpm}. To study the impact of model size, we test the 3B, 7B, 32B, and 72B versions of Qwen2.5-VL, as well as the 4B, 8B, 26B, 38B, and 78B  versions of InternVL2.5. We also evaluate pure language models (e.g., LLaMA-3.1 \cite{grattafiori2024llama}, Qwen2.5 \cite{yang2024qwen2}) by replacing image inputs with textual descriptions.

We report results under three different settings: \textit{original results}, \textit{oracle results}, and \textit{KB results}. The \textit{original results} correspond to answers generated directly based on the image and question. The \textit{oracle results} are obtained by providing additional evidence at varying granularities. The \textit{KB results} involve supplying a knowledge base, from which the model retrieves relevant evidence using a retrieval method for reference.






\subsection{Original Results}
Table~\ref{model_performance_original} presents the original experimental results, in which only the question and the image are provided as input to the MLLMs. We draw the following conclusions: M$^3$-VQA presents a significant challenge to current MLLMs. Even the best-performing model, Qwen2.5-VL-72B-Instruct, achieves only 32.6\% accuracy. Larger model sizes improve performance. Comparing the 3B, 7B, 32B, and 72B versions of Qwen2.5-VL, as well as the 4B, 8B, 26B, 38B, and 78B versions of InternVL2.5, we observe consistent performance gains as parameter count increases. Pure language models perform worse than vision-language models across all settings. This indicates that image descriptions alone are insufficient to capture all visual information, especially with fine-grained entities.

To further validate the multimodal nature of M$^3$-VQA, we also test the model's performance when the image is removed (only the textual question is given, abbr. \textit{Q-Only}). As shown in Table~\ref{model_performance}, a 8.9\% gap between \textit{Original} and \textit{Q-Only} shows that image input is essential. The 15.6\% accuracy of \textit{Q-Only} is partly due to correct answers based on text entity properties and our IoU-based metric, which gives partial credit. Models may also guess correctly on factual or counting questions.


\subsection{Oracle Results}

In the oracle setting, the model is provided with a pre-annotated golden evidence chain. This setting primarily evaluates the model’s ability to perform reasoning using multi-hop evidence. We categorize the evidence into three granularities: \textit{Sentence}, which corresponds to the most precise sentence in the linked Wikipedia page which supports the answer; \textit{Section}, referring to the Wikipedia section containing the evidence; and \textit{Entity-Name}, indicating the entity name associated with the evidence.

As shown in Table~\ref{model_performance}, even under the best-case \textit{Sentence} setting, the top-performing model achieves only 58.7\% accuracy, with an average of 49.9\%. This highlights the difficulty current models have with multi-hop and multi-entity reasoning. The 10.2\% difference between \textit{Entity-Name} and \textit{Original} setting suggests that identifying fine-grained visual entities remains a major hurdle. Adding visual entity recognition shows great promise. Comparing \textit{Sentence} and \textit{Entity-Name} (both having access to entity identity), there's still a 15.2\% gap, showing that recalling detailed attributes remains challenging. Retrieval helps significantly here. There’s a 4.8\% accuracy gap between \textit{Sentence} and \textit{Section}, showing that providing more precise local information boosts performance. Smaller supporting documents lead to more verifiable and explainable answers.

\begin{table*}[t]
  \vspace{-13pt}
  \centering
  \small
  \begin{tabular}{lccccc}
\toprule
 \multirow{2}{*}{\textbf{Model}} 
& \multicolumn{3}{c}{\textbf{Evidence Provided}} 
& \multicolumn{2}{c}{\textbf{No Evidence}}  \\
\cmidrule(lr){2-4}
\cmidrule(lr){5-6}
& \textbf{Sentence}
& \textbf{Section}
& \textbf{Entity-Name} 
& \textbf{Original} 
& \textbf{Q-Only} \\
\midrule 

Qwen2.5-VL-7B-Instruct \cite{bai2025qwen2} & 49.0 & 42.7 & 31.8 & 22.5 & 14.4  \\
Qwen2.5-VL-72B-Instruct \cite{bai2025qwen2} & 58.4 & 52.4 & \textbf{45.1} & \textbf{32.6} & 16.1  \\
InternVL2.5-8B \cite{chen2024expanding}  & 46.8 & 42.1 & 25.7 & 18.7 & 11.8 \\
InternVL2.5-78B \cite{chen2024expanding}  & \textbf{58.7} & \textbf{55.2} & 45.0 & 31.3 & 14.7 \\
LLaVA-OneVision-7B \cite{li2024llava} & 47.2 & 43.6 & 31.7 & 22.0 & 17.3 \\
MiniCPM-V-2.6 (8B) \cite{yao2024minicpm} & 46.9 & 39.1 & 33.6 & 23.5  & \textbf{18.5} \\
Llama-3.1-8B-Instruct (\textit{text}) \cite{grattafiori2024llama}  & 42.3 & 40.5 & 29.7 & 20.6 & 16.1 \\
\midrule
\textit{Average} & 49.9 & 45.1 & 34.7 & 24.5 & 15.6 \\
\bottomrule
\end{tabular}
\vspace{-5pt}
  \caption{Performance comparison of various models under different settings.}
  \label{model_performance}
\end{table*}

\begin{table*}[t]

  \centering
  \small
\begin{tabular}{lccc}
\toprule
\textbf{Model}
&  \textbf{Original}
&  \textbf{Heuristic Retrieval}
&   \textbf{Agentic Retrieval} \\
\midrule

 Qwen2.5-VL-7B-Instruct \cite{bai2025qwen2}      &  22.5  & 25.7 & 31.2 \\
 Qwen2.5-VL-72B-Instruct \cite{bai2025qwen2}     &  \textbf{32.6}  &  \textbf{36.6}  & \textbf{38.9}\\
 InternVL2.5-8B \cite{chen2024expanding}               &  18.7  & 25.3  & 29.4 \\
 InternVL2.5-78B	\cite{chen2024expanding}              &  31.3  & 33.2  & 36.4 \\
 LLaVA-OneVision-7B  \cite{li2024llava}          &  22.0  &  25.8 & 28.9 \\
 MiniCPM-V-2.6 (8B)   \cite{yao2024minicpm}         &  23.5  &  24.7 & 28.0 \\
Llama-3.1-8B-Instruct (\textit{text}) \cite{grattafiori2024llama}  & 20.6  &  23.4 & 28.3 \\

\bottomrule
\end{tabular}
\vspace{-5pt}
  \caption{Performance comparison of various models under KB setting.}
  \label{model_performance_rag}
  \vspace{-10pt}
\end{table*}

\subsection{KB Results}

In the knowledge-base (KB) setting, the model has access to a manually constructed knowledge base, in addition to the question and image. This setting allows us to evaluate the model’s ability to perform multi-hop reasoning through information retrieval. We design two general retrieval algorithms to assist the tested MLLM in answering questions: \textit{Heuristic Retrieval} and \textit{Agentic Retrieval}.

\textbf{Heuristic Retrieval.} We implement a two-stage heuristic retrieval: \textit{(1) Text Retrieval:} The question is used as a query to retrieve the most relevant text passages from the KB; \textit{(2) Image Retrieval:} The image is used to retrieve similar visual entities from the KB. Once identified, the question is used to search within the corresponding Wikipedia articles for relevant text. The evidence retrieved from both text and image search is concatenated and passed as context to the model. 

\textbf{Agentic Retrieval.}
We developed agentic retrieval model composed of Planner, Executor, and Solver. The Executor includes tools such as object detection, text retrieval, and image retrieval. The Planner is responsible for devising the problem-solving steps and invoking the relevant tools. It first calls the object detection module within the Executor to detect and segment the image. Then, it either uses the segmented image to perform image retrieval or generates single-hop queries for text retrieval. The Executor carries out the actual execution of these tool calls. Finally, the Solver produces the answer based on the retrieved information.

For heuristic retrieval and agentic retrieval, the tested MLLMs are used as the Solver; GPT-4o is used as the Planner; Qwen2.5-VL-7B-Instruct for object detection; BGE-Large-en-v1.5 \cite{xiao2024c} for text embedding; and CLIP-ViT-Large \cite{radford2021learning} for image embedding. Table~\ref{model_performance_rag} presents the experimental results. We draw the following conclusions:

\textit{Heuristic retrieval} methods suffer from overloaded queries. Using the full question as a single text query is problematic when the question contains multiple entities or sequential sub-questions. Similarly, using the entire image may hinder visual retrieval, especially when multiple entities are present. These one-shot strategies place excessive burden on a single query, often retrieving superficially relevant but unhelpful content. \textit{Agentic retrieval} significantly outperforms all other models, including both open-source and closed-source MLLMs and LLM using heuristic retrieval. We attribute this to two factors: agentic retrieval decomposes complex questions into simpler sub-questions, reducing the burden of single-step retrieval; it rethinks both the content retrieved and the sub-questions, ensuring sufficient information is gathered.

M$^3$-VQA presents significant challenges to existing models in terms of multi-hop and multi-entity retrieval. A 20\% gap between \textit{agentic retrieval} and \textit{Sentence} indicates the challenge of accurate retrieval. Accuracy with retrieved evidence generally falls between the \textit{Original} and \textit{Oracle} settings. This confirms the practical effectiveness of multimodal, multi-hop, and multi-entity retrieval. The dataset leaves ample room for future research. 

\subsection{Impact of Hop and Entity Counts}

\begin{figure*}[t]
  \centering
  \vspace{-13pt}
  \includegraphics[width=0.71\textwidth]{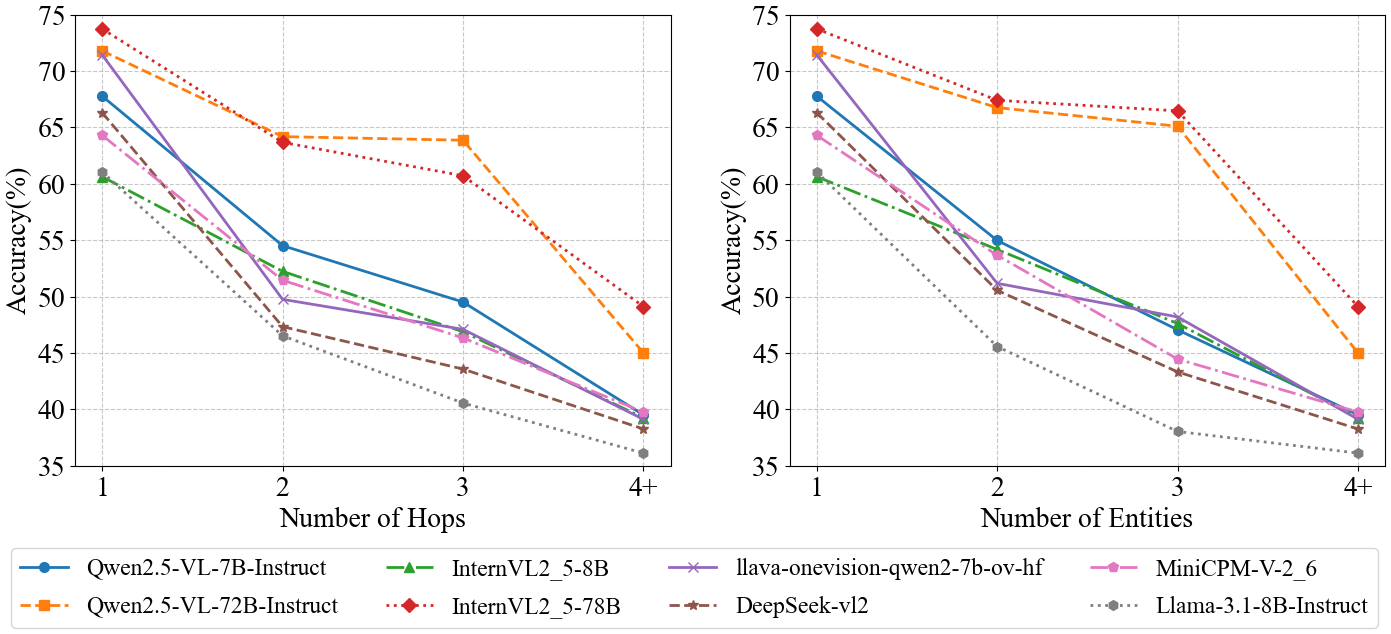}
  \vspace{-5pt}
  \caption{Model accuracy across different hop and entity counts, under the sentence evidence setting.}
  \label{hop_entity}
\end{figure*}

\begin{table*}[t]
  \centering
  \resizebox{\linewidth}{!}{
  \small
  \begin{tabular}{clcccccc}
\toprule
\textbf{Question}
& \textbf{\multirow{2}{*}{Model}}
& \textbf{No}
& \multicolumn{3}{c}{\textbf{Partial Evidence}} 
& \textbf{Full} \\
\textbf{Type}
& 
& \textbf{Evidence}
& \textbf{1}
& \textbf{2}
& \textbf{3}
& \textbf{Evidence} \\
\midrule 
\multirow{5}{*}{\makecell[c]{Parallel Reasoning\\(4 Entities)}}
& Qwen2.5-VL-7B \cite{bai2025qwen2}        & 21.7 & 28.1 & 34.8  & 36.7 & 39.5 \\
& Qwen2.5-VL-72B  \cite{bai2025qwen2}      & \textbf{29.2} & \textbf{34.0} & \textbf{38.6}  & \textbf{39.5} & \textbf{45.0} \\ 
& InternVL2.5-8B  \cite{chen2024expanding}      & 17.4 & 27.9 & 33.4  & 36.6 & 39.3 \\
& MiniCPM-V-2.6  \cite{yao2024minicpm}       & 25.1 & 31.8 & 36.1  & 37.4 & 39.8 \\
& Llama-3.1-8B-Instruct (\textit{text}) \cite{grattafiori2024llama}  & 17.4 & 23.0 & 26.7  & 31.7 & 36.1 \\
\midrule 
\multirow{5}{*}{\makecell[c]{Sequential Reasoning\\(3 Hops)}}
& Qwen2.5-VL-7B \cite{bai2025qwen2}        & 25.9 & 33.8 & 43.7  & - & 59.0 \\ 
& Qwen2.5-VL-72B \cite{bai2025qwen2}       & \textbf{26.2} & \textbf{35.3} & \textbf{45.7}  & - & \textbf{59.5} \\
& InternVL2.5-8B  \cite{chen2024expanding}      & 17.6 & 29.0 & 36.9  & - & 44.0 \\
& MiniCPM-V-2.6    \cite{yao2024minicpm}     & 13.4 & 29.6 & 39.5  & - & 53.9 \\
& Llama-3.1-8B-Instruct (\textit{text}) \cite{grattafiori2024llama}  & 20.5 & 31.0 & 41.2  & - & 50.7 \\
\bottomrule
\end{tabular}
}
  \caption{Model accuracy under varying degrees of gold reasoning evidence coverage. 
  }
  \vspace{-5pt}
  \label{model_performance_evidence}
\end{table*}

To understand how question complexity affects model performance, we analyze accuracy as a function of hop count and entity count, using line charts under the \textit{Oracle setting}, where full reasoning support is provided. As shown in Figure~\ref{hop_entity}, model accuracy monotonically decreases with more reasoning hops or more involved entities. This confirms that increased logical complexity poses greater challenges for current models, even when relevant evidence is explicitly given.

However, this performance drop becomes less pronounced under less supportive conditions, such as the \textit{Original setting}, where no additional external evidence is provided. In these cases, models sometimes perform better on higher-hop or multi-entity questions. This phenomenon is primarily attributed to the structure and content of the questions themselves.
Specifically, higher hop or entity complexity tends to introduce more textual entities into the question, and these entity names are explicitly mentioned in the question itself. These elements often relate to general world knowledge or language-based associations that models can answer without full reasoning or retrieval. In contrast, questions focused on visual entities typically require grounded image understanding, which is more challenging in the absence of evidence.

\subsection{Impact of Reasoning Support Coverage}

To investigate how partial evidence affects performance on complex reasoning tasks, we design a controlled evaluation where all questions have fixed complexity: either 4-entity parallel reasoning or 3-hop sequential reasoning. For each question, we vary the number of reasoning steps for which gold supporting evidence is provided—from none to full coverage. This setup simulates different levels of difficulty by controlling how much of the reasoning chain is revealed to the model.

As shown in Table~\ref{model_performance_evidence}, model accuracy improves consistently as more evidence is made available. For both reasoning paradigms, even partial support (e.g., 1 or 2 hops/entities) leads to large performance gains over the no-evidence baseline. The most substantial improvement occurs when near-complete or full evidence is given, especially in sequential reasoning tasks where intermediate steps are critical. These results highlight the strong dependency of multi-entity and multi-hop VQA models on the availability of high-quality intermediate knowledge, and motivate future work on retrieval and grounding strategies.

We strongly recommend reading Appendix~\ref{appx:H} for additional experimental results.

\section{Conclusion}
In this work, we introduce M$^3$-VQA, a challenging and novel benchmark designed to push the limits of MLLMs in fine-grained entity understanding and complex multi-hop reasoning across multimodal inputs. Unlike prior VQA datasets that focus on simpler, single-entity questions and coarse-grained categories, M$^3$-VQA presents diverse multi-entity scenarios requiring both sequential and parallel reasoning over multiple documents and modalities. Our comprehensive evaluation of 16 leading MLLMs under varying conditions (without evidence, with gold evidence, and with retrieval-augmented input) reveals substantial gaps in current model capabilities. The poor performance without external information highlights the critical need for effective knowledge integration. Moreover, the significant gains achieved through precise evidence and reasoning-aware agentic retrieval demonstrate the vital role of structured, traceable reasoning and sophisticated retrieval mechanisms for complex multimodal understanding. We believe M$^3$-VQA sets a new, rigorous standard for evaluating and advancing the reasoning abilities of future MLLMs, encouraging further research into more robust knowledge acquisition and multi-hop reasoning strategies.



\section*{Acknowledgements}
This research is supported by Artificial Intelligence-National Science and Technology Major Project (2023ZD0121200), and the National Natural Science Foundation of China (62437001, 62436001, 62531026), the Key Research and Development Program of Jiangsu Province under Grant BE2023016-3, and the Natural Science Foundation of Jiangsu Province under Grant BK20243051.

\section*{Limitations}

Our dataset has several limitations that should be acknowledged. First, it is currently limited to English-language content, which restricts its applicability to multilingual or cross-lingual research settings. Future work could extend the dataset to include other languages supported by Wikipedia and beyond, facilitating broader cultural and linguistic coverage. Second, the dataset focuses primarily on knowledge derived from Wikipedia. While Wikipedia is a rich and widely-used source, it may not comprehensively represent specialized domains. Future expansions could explore additional domains such as biomedical literature, legal texts, or cultural artifacts. Third, although we employed a combination of automatic generation and manual filtering for question construction, the presence of a small number of errors or ambiguous instances is inevitable. We encourage users to be mindful of this and consider further validation in downstream tasks. Additionally, while we evaluate a wide range of models under multiple settings, our current study does not systematically assess different reasoning strategies, such as chain-of-thought (CoT) \cite{wei2022chain} prompting. Exploring the impact of such reasoning techniques remains an important direction for future work.

\section*{Ethical Considerations}

\subsection*{Data Sources and Usage Permission}
All images used in this dataset are sourced from publicly licensed datasets or from publicly accessible resources explicitly authorized by the rights holders. We strictly adhere to the usage agreements and licensing terms of each data source, and all data is used solely for academic research and non-commercial purposes.

\subsection*{Personal Privacy and Identifiable Information}
We acknowledge that images may contain personally identifiable information, particularly facial features and license plates. To address this, we have blurred license plates to prevent vehicle identification and misuse. While the dataset includes some images of public figures, all such images are publicly available on authorized platforms, and their use complies with fair use principles and the terms of the original datasets. We have not collected or published any unauthorized private photos or images.

\subsection*{Bias and Fairness}
Although we have made efforts to ensure diversity in image content and question-answer pairs, the dataset may still contain cultural or regional biases inherited from the original image collections. We encourage downstream users to carefully examine and mitigate potential bias impacts when using this dataset for model training or evaluation. Additionally, our work employs pretrained large language models for generating, analyzing, or evaluating VQA content. These models may carry inherent biases from their training data, including linguistic, cultural, or political tendencies. Such issues are beyond our control, and we advise readers to interpret results with caution.

\subsection*{Intended Use and Potential Misuse}
This dataset is intended to advance academic research in multimodal AI systems, particularly in the development and evaluation of visual question answering tasks. We explicitly oppose any use of this dataset for privacy infringement or discriminatory technology development. We strongly recommend that users adhere to relevant ethical guidelines and legal regulations during data usage.

\subsection*{Environmental Impact}
LLMs were used during our research, which operate on high-performance computing hardware and require significant energy resources, potentially contributing to carbon emissions. We encourage researchers to adopt efficient resource management practices to reduce carbon footprints during model development and evaluation.


\bibliography{custom}

\appendix

\section{Additional Related Work}
\label{appx:A}
In addition to the related work discussed in Section 2, there are several more relevant areas:

\subsection{Visual Question Answering}
Visual Question Answering refers to answering questions about images. Early VQA datasets such as DAQUAR \cite{malinowski2014multi}, GQA \cite{hudson2019gqa}, VQAv1 \cite{antol2015vqa}, VQAv2 \cite{goyal2017making}, CLEVR \cite{johnson2017clevr}, FM-IQA \cite{gao2015you}, and Visual Genome \cite{krishna2017visual} mainly focus on reasoning based on image content, without requiring inference over named entities in the image or external knowledge.

\subsection{Multi-hop Reasoning}
There is a wealth of multi-hop textual datasets in the field of natural language processing \cite{jiang2020hover, yang2018hotpotqa, talmor2018web, joshi2017triviaqa, khot2020qasc, dunn2017searchqa, mihaylov2018can, aly2021feverous}, covering tasks such as question answering and fact verification. Notably, HOTPOTQA \cite{yang2018hotpotqa} introduced the concept of "bridge entities" to construct multi-hop QA datasets. For example, in the question, “When was the singer and songwriter of Radiohead born?”, one must first infer that the singer and songwriter of Radiohead is Thom Yorke, and then retrieve his birth date. Here, Thom Yorke serves as the bridge entity. We adopt this concept in constructing our sequential multi-hop questions.

\subsection{Entity Recognition}
Fine-grained visual entity recognition is a key challenge in M$^3$-VQA. Accurate reasoning and correct answers rely on correctly identifying the names of entities within images. Early work on fine-grained recognition focused on narrow domains such as plant identification \cite{van2018inaturalist}, flowers \cite{nilsback2008automated}, bird species \cite{wah2011caltech}, dogs \cite{khosla2011novel}, cats \cite{parkhi2012cats}, fish species \cite{kay2021fishnet}, food \cite{bossard2014food}, cars \cite{krause20133d}, or aircraft types \cite{maji2013fine}—typically relying on deep learning models for classification. Recently, OVEN \cite{hu2023open} unified labels from these datasets under Wikipedia’s 6 million entities, proposing the task of Open-domain Visual Entity Recognition (OVEN), which requires models to link images to Wikipedia entities based on textual queries. These queries can be viewed as simple VQA tasks without detailed attribute reasoning. Unfortunately, OVEN lacks images involving multiple entities, which led us to build our own dataset from scratch.

\subsection{Automatic Question Generation}
Some prior work \cite{changpinyo2022all, mensink2023encyclopedic, chen2023can, shah2019kvqa, li2024synthesize} has explored the automatic generation of single-hop VQA datasets. These approaches can be broadly categorized into three types. The first \cite{chen2023can, shah2019kvqa} uses question templates combined with SPARQL queries to Wikidata for retrieving answers. This method ensures high accuracy but suffers from limited diversity. The second approach \cite{changpinyo2022all, mensink2023encyclopedic} employs trained models to generate questions from free text, offering some diversity, though answers often depend heavily on the specific text. The third \cite{li2024synthesize} uses large language models (LLMs), which can automatically generate questions. However, in our tests, neither trained models nor LLMs were able to generate high-accuracy multi-hop datasets. Therefore, we used a Wikidata-based method to construct multi-hop, multi-entity datasets. We further employed LLMs for question rewriting, combined with strict filtering and sampling to enhance question diversity.

\section{Proof of Theorem}
For each $(I, Q, A)$ triplet, we define: $E_I = \{e^{I}_{1}, e^{I}_{2}, \ldots\}$ as the set of visual entities in the image; $E_Q = \{e^{Q}_{1}, e^{Q}_{2}, \ldots\}$ as the set of textual entities in the question. Complexity definitions are as follows: 
\paragraph{Image Parallel Complexity} Defined as the number of visual entities in image $I$, denoted as $IP = |E_I|$. 
\paragraph{Text Parallel Complexity} Defined as the number of explicitly mentioned textual entities in question $Q$, denoted as $TP = |E_Q|$. 
\paragraph{Sequential Complexity} The question $Q$ can be represented as a sequence $Q = (q_1, q_2, \ldots, q_s)$, where each $q_i$ for $i = 2, 3, \ldots, s$ depends on the answer to $q_{i-1}$. The maximum length $s$ is defined as the sequential complexity: $S = \max(s)$. 
\paragraph{Entity Complexity} The total number of visual and textual entities, defined as $P = IP + TP$. 
\paragraph{Hop} The number of Wikipedia pages needed to answer question $Q$.

\paragraph{Theorem}$Hop = IP + TP + S - 1$.

$Proof$: Each Wikipedia page corresponds to one entity. Considering the maximum number $s$, the questions $q_2, \ldots, q_s$ each introduce a new entity (i.e., the answer to the previous question), while the entity in $q_1$ can only be from existing ones, i.e., $E_{q_1} = E_I \cup E_Q$. Therefore:
\[
\begin{aligned}
Hop &= |E_I \cup E_Q \cup E_S| \\
&= |E_I \cup E_Q \cup E_{q_2} \cup \ldots \cup E_{q_s}| \\
&= |E_I| + |E_Q| + |E_{q_2}| + \ldots + |E_{q_s}| \\
&= IP + TP + (s - 1) \\
&= IP + TP + S - 1
\end{aligned}
\]

\section{Extending M$^3$-VQA to Other Tasks}

The M$^3$-VQA dataset is not only suitable for visual question answering (VQA) tasks, but it also naturally supports or can be extended to the following tasks:

\subsection{Multimodal Retrieval}

Based on M$^3$-VQA, multimodal retrieval involves multi-hop and multi-entity retrieval, termed as \textbf{M3RAG} (Multimodal Multi-hop Multi-entity Retrieval-Augmented Generation). This is also a novel task that has not been previously explored. The M3RAG task can be represented as a function $f: (I, Q, K) \rightarrow G$, where:

\begin{itemize}
    \item $I$ is an image. Each image $I \in \mathcal{I}$ is represented by pixels and may contain multiple visual entities;
    \item $Q$ is a question. Each question $Q \in \mathcal{Q}$ is represented by text and may contain multiple textual entities and sub-questions;
    \item $K$ is an external Wikipedia knowledge base, consisting of free-form text and images;
    \item $G$ is the gold evidence, with each $G \in \mathcal{G}$ corresponding to a subset of pages in $K$.
\end{itemize}

We define $IP$, $TP$, and $S$ similarly to M$^3$-VQA. $Hop$ is defined as the total number of knowledge pages that the gold evidence $G$ belongs to. In our dataset, we have:

\[
Hop = IP + TP + S - 1
\]

\subsection{Multi-hop Reasoning}

When evidence retrieval is not required and gold evidence is directly provided, M$^3$-VQA reduces to the \textbf{M3REASONING} task. In this task, the model bypasses the retrieval challenge and focuses on performing multi-step reasoning based on the given knowledge to answer the question. The M3REASONING task is defined as a function $f: (I, Q, G) \rightarrow A$, where:

\begin{itemize}
    \item $I$ is an image. Each image $I \in \mathcal{I}$ is represented by pixels and may contain multiple visual entities;
    \item $Q$ is a question. Each question $Q \in \mathcal{Q}$ is represented by text and may contain multiple textual entities and sub-questions;
    \item $G$ is the gold evidence, with each $G \in \mathcal{G}$ containing several evidence items;
    \item $A$ is the answer, with each $A \in \mathcal{A}$ being a string, time, number, etc.
\end{itemize}

We define $IP$, $TP$, and $S$ similarly to M$^3$-VQA. $Hop$ corresponds to the number of reasoning steps. In our dataset, we have:

\[
Hop = IP + TP + S - 1
\]

\subsection{Visual Entity Recognition}

By discarding the textual questions in M$^3$-VQA, we can instead construct a textual query $Q$, such as ``What kind of lion is on the left side of the picture?''. Additionally, we construct an entity label set $E$ based on Wikipedia. This leads to the task of \textbf{Open-domain Multiple Visual Entity Recognition (OMVER)}, which is also a previously unexplored task. The OMVER task is defined as a function $f: (I, Q, K) \rightarrow A$, where:

\begin{itemize}
    \item $I$ is an image. Each image $I \in \mathcal{I}$ is represented by pixels and may contain multiple visual entities;
    \item $Q$ is a query. Each query $Q \in \mathcal{Q}$ is represented by text and indicates the intention of identifying elements in image $I$;
    \item $K$ is a knowledge base, represented as $K = \{(e, p(e), t(e)) \mid e \in E\}$, a set of triples. Each triple contains an entity $t(e)$ (i.e., the entity's name, description, etc.) and a (possibly empty) set of associated images $p(e)$;
    \item $A$ is the answer space, where each $A \in \mathcal{A}$ is an entity from $E$.
\end{itemize}

\subsection{Summary}

Note that these three tasks are in fact sub-tasks of M$^3$-VQA, but they can also be studied independently. On the other hand, improving M$^3$-VQA often requires advances in visual entity recognition, multimodal retrieval, and multi-hop reasoning. Progress in these three tasks will significantly promote the development of M$^3$-VQA. Furthermore, merging visual entity recognition and multimodal retrieval into a single unified step may be a promising future direction.

\section{Dataset Construction Details}
\label{appx:D}
\subsection{Image Sources}

\paragraph{Manual Collection}
To gather more fine-grained, multi-entity images, we used search engines like Google and Baidu. Specifically, annotators crafted queries likely to yield multi-entity images—for example, “Ragdoll cat and Maine Coon.” They then searched for images and manually reviewed and filtered the results.

\paragraph{CelebTo} \cite{zhong2018compact}. This dataset  contains images with multiple labeled celebrities in each picture. It can be used as a benchmark for evaluating the retrieval of a set of identities. The dataset consists of 194,000 images, totaling 546,000 faces, covering 2,622 labeled celebrities. 59\% of the faces correspond to these 2,622 celebrities, while the remaining faces are considered "unknown" individuals. The images in this dataset are sourced from Google Image Search and validated through manual labeling.

\paragraph{FGVD} \cite{khoba2022fine}. It is a dataset for fine-grained vehicle detection, with images collected from mobile cameras mounted on cars. The FGVD dataset is challenging because the vehicles appear in complex traffic scenes with a diversity of types, scales, poses, occlusions, and lighting conditions within and across categories. The dataset contains 5,502 scene images and covers 210 unique fine-grained labels, including various vehicle types, and is organized into a three-layer hierarchical structure. While previous classification datasets have included manufacturer information for different vehicle models, FGVD introduces new category labels for classifying two-wheeled vehicles, auto rickshaws (three-wheelers), and trucks.

\paragraph{FlickrLogos-47} \cite{romberg2011scalable}. The dataset  contains photos showcasing brand logos and is designed for evaluating logo detection and recognition systems in real-world images. This dataset is constructed from the same images as the FlickrLogos-32 dataset but has been re-annotated to fix missing labels and add more categories.

\paragraph{LogosInTheWild} \cite{tuzko2017open}. This dataset  consists of web images and their corresponding logo annotations, collected via Google Image Search.

\paragraph{Menu-match} \cite{beijbom2015menu}. The dataset includes meal images from three restaurants: an Asian restaurant offering a buffet-style service where customers can choose 1 to 3 side dishes served with brown or white rice; an Italian restaurant serving a variety of pizzas, lasagna, pasta, and accompanying breadsticks or salads; and a soup restaurant providing 10 types of soups with a choice of 5 types of bread. The dataset contains 646 images, annotated with 1,386 food items, covering 41 categories.

\paragraph{Oktoberfest} \cite{ziller2019oktoberfest}. It is a real, diverse, and challenging dataset for object detection in images. The data was collected from a beer tent scene in Germany and includes 15 different categories of food and drink items.

\paragraph{UEC-FoodPix} \cite{okamoto2021uec}.  It is a food image dataset with segmentation masks, containing 9,000 training images and 1,000 test images. The segmentation masks are enhanced according to food categories. In UECFoodPix, the mask images are generated using bounding boxes and the GrabCut method. The mask images have pixel-level labels for 103 food categories, which are stored only in the red (R) channel of the images.

\paragraph{UNIMIB2016} \cite{ciocca2016food}. This dataset  is used for food recognition and segmentation tasks. It contains 1,027 images of trays with multiple food items, covering 73 food categories. All images have been manually labeled using polygons for each food instance, with corresponding food labels.

\paragraph{Infoseek} \cite{chen2023can}. This  is a visual question answering dataset focused on information retrieval-type questions that cannot be answered through common sense knowledge. It collects high-quality visual information retrieval question-answer pairs through multi-stage manual annotation. Additionally, it constructs a large-scale automatically collected dataset by combining existing visual entity recognition datasets with Wikidata, providing over one million examples for model fine-tuning and validation.

\paragraph{EVQA} \cite{mensink2023encyclopedic}. This  is a large-scale visual question answering (VQA) dataset focused on visual questions about fine-grained categories and detailed attributes of instances. The dataset contains 221,000 unique question-answer pairs, each paired with (up to) 5 images, forming 1 million VQA samples. It provides a controlled knowledge base derived from Wikipedia and offers supporting evidence for each answer.

\subsection{Entity Linking}

\paragraph{Exact Matching Stage}
We used QLever \cite{bast2017qlever}, a highly efficient SPARQL engine capable of handling large-scale knowledge graphs like the full Wikidata \cite{vrandevcic2014wikidata} dump (dated August 22, 2022). We queried the Wikidata QIDs using the original label texts. If the query returned a unique result, we accepted it.

\paragraph{Manual Matching Stage}
If the query yielded multiple results or if poor linking was detected, annotators manually resolved the match. We excluded entities that could not be found in Wikidata or lacked an English Wikipedia page.

\subsection{Knowledge Base}

To align with Wikidata, we selected the English Wikipedia \cite{burns2023suite} snapshot closest in time (August 13, 2022), containing approximately 2 million articles. We retained the pages of all entities included in M$^3$-VQA. For each entity, we sampled about 5 hard negative examples. Specifically, we used all-mpnet-base-v2 to compute embeddings for Wikipedia article summaries and retrieved the top 20 most similar entities using FAISS, randomly selecting 5 as hard negatives. The candidate entity set included all M$^3$-VQA entities and their hard negatives. From each entity’s Wikipedia page, we selected at least 3 sections—including the one containing golden evidence—to form the text corpus. The first image of each entity page was used to build the image corpus. These two corpora together formed the knowledge base. Importantly, M$^3$-VQA requires retrieval from the full constructed knowledge base, not just from the positive and hard negative examples. This strategy avoids searching over all 2 million articles (reducing computational cost and evaluation time) while maintaining retrieval difficulty through the inclusion of challenging negative samples.

\subsection{Quality Control}

\paragraph{Annotator Selection}
To ensure annotation quality, we provided annotators with thorough training. They attended in-person tutorial sessions covering annotation guidelines and common mistakes. Then, they took a qualification test; those who failed received personalized feedback and were retested. Only those who passed could participate in the main task. All annotators were at least undergraduate students. We explicitly explained to them how the collected data would be utilized. This project is voluntary, and all collaborators were informed of the authorship mechanism before joining.

\paragraph{Filtering}
We applied strict filtering at multiple stages:

\begin{itemize}

\item Low-quality images were removed.

\item Entities not found in Wikidata or Wikipedia were discarded.

\item Questions lacking corresponding entity attributes in Wikidata were eliminated.

\item Sequential multi-hop questions were verified with a large language model (LLM): if the new question’s answer didn’t match the original, or if it was too short/long, it was discarded.

\item For IQA (Image-Question-Answer) pairs, since Wikidata and Wikipedia are independently crowd-sourced, some answers from Wikidata might not appear in Wikipedia. We removed any question whose answer couldn’t be found in the relevant Wikipedia page.

\item Annotators also filtered out poorly structured questions.

\end{itemize}

In total, we retained approximately 550,000 IQA pairs.

\paragraph{Sampling}
To ensure balance across different types and complexities, and to maintain diversity in questions, entities, and images, we performed subsampling on the original dataset. Details are in Appendix E.

\paragraph{Quality Evaluation}
To further assess quality, we evaluated annotation accuracy. Experts reviewed 200 stratified samples from M$^3$-VQA. For each sample, they determined the answer based on the image, question, Wikipedia page, and evidence, and compared it with the provided annotation. Results showed that 93\% of expert answers matched the annotations. For comparison, A-OKVQA has 86\% correctness, INFOSEEK 95\%, and EVQA 86\%. Thus, M$^3$-VQA demonstrates high annotation quality.

\section{Additional Dataset Statistics}
\label{appx:E}
\subsection{Complexity}

\begin{table}
\centering
\begin{tabular}{lccccc}
      \toprule
      \textbf{}
        & \textbf{S} 
        & \textbf{P} 
        & \textbf{TP} 
        & \textbf{IP} 
        & \textbf{hop} \\
      \midrule
        0 & - & - & 6929 & - & -\\
        1 & 11225 & 2992 & 1909 & 6992 & 1092\\
        2 & 998 & 2183 & 2287 & 2827 & 3181\\
        3 & 902 & 3644 & 1000 & 2355 & 4546\\
        4+ & - & 4306 & 1000 & 951 & 4306\\
      \bottomrule
    \end{tabular}
\caption{Complexity of M$^3$-VQA.}
\label{dataset_complexity}
\end{table}

\begin{figure*}
  \centering
  \includegraphics[width=0.8\textwidth]{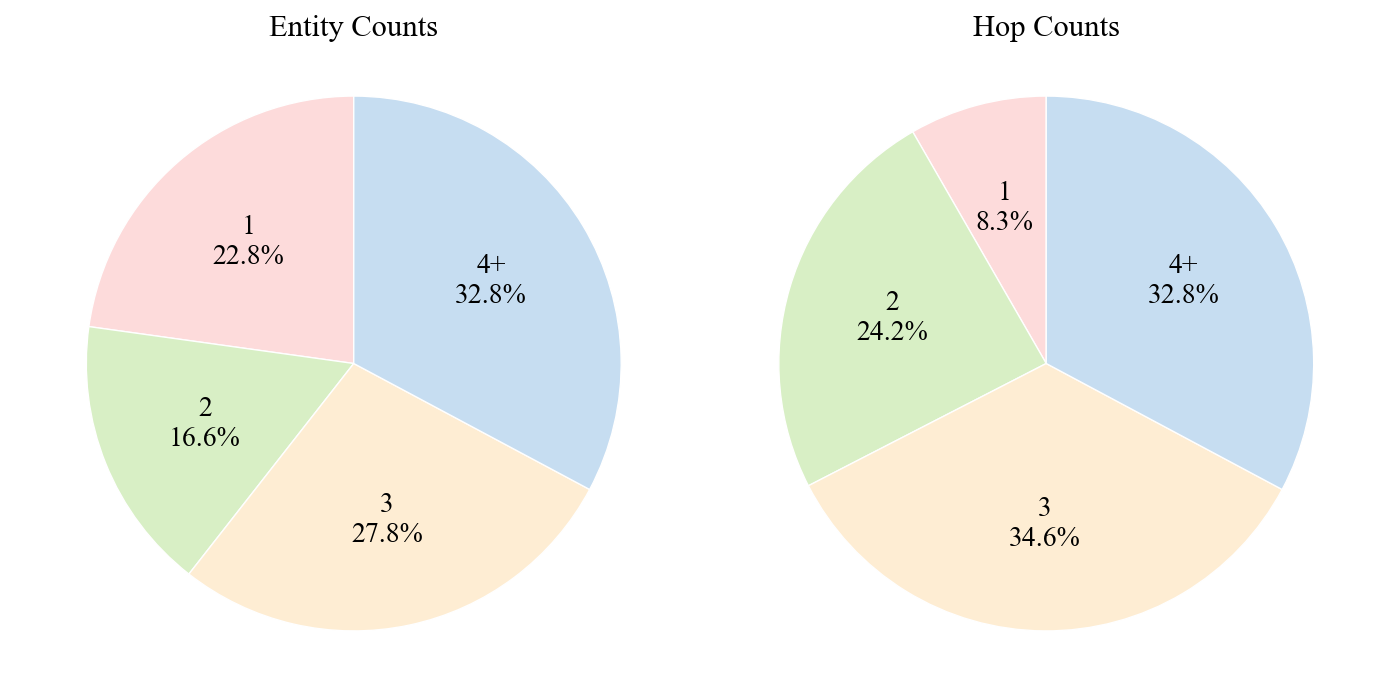}
  \caption{Complexity of M$^3$-VQA.}
  \label{count}
\end{figure*}

Table~\ref{dataset_complexity} and Figure~\ref{count} presents the specific counts of questions with varying levels of complexity. In terms of hop count, the dataset includes 1,092 single-hop questions, 3,181 two-hop questions, 4,546 three-hop questions, and 4,306 questions with four or more hops. In terms of the number of entities, the dataset includes 2,992 single-entity questions, 2,183 two-entity questions, 3,644 three-entity questions, and 4,306 questions with four or more entities.

\subsection{Detailed Complexity of M$^3$-VQA}

\begin{table}
\centering
\begin{tabular}{cccccc}
\toprule
\diagbox{\textbf{TP}}{\textbf{IP}} & 1 & 2 & 3 & 4 & 5+ \\
\midrule
0 & 1092 & 1183 & 2000 & 500 & 254 \\
1 & 1000 &  644 &  171 &  73 &  21 \\
2 & 1000 & 1000 &  184 &  75 &  28 \\
3 & 1000 &   -  &   -  & -   &  -  \\
4 & 1000 &   -  &   -  &  -  &   - \\
\bottomrule
\end{tabular}
\caption{Distribution of $IP$ and $TP$ under $S = 1$}
\label{tp_ip}
\end{table}

In our dataset, there are 998 questions with $S = 2, IP = 1, TP = 0$, and 902 questions with $S = 3, IP = 1, TP = 0$. All other questions have $S = 1$. Under the condition of $S = 1$, the distribution of $IP$ and $TP$ in the dataset is shown in Table~\ref{tp_ip}.

\subsection{Question Type}
\begin{figure*}
  \centering
  \includegraphics[width=0.8\textwidth]{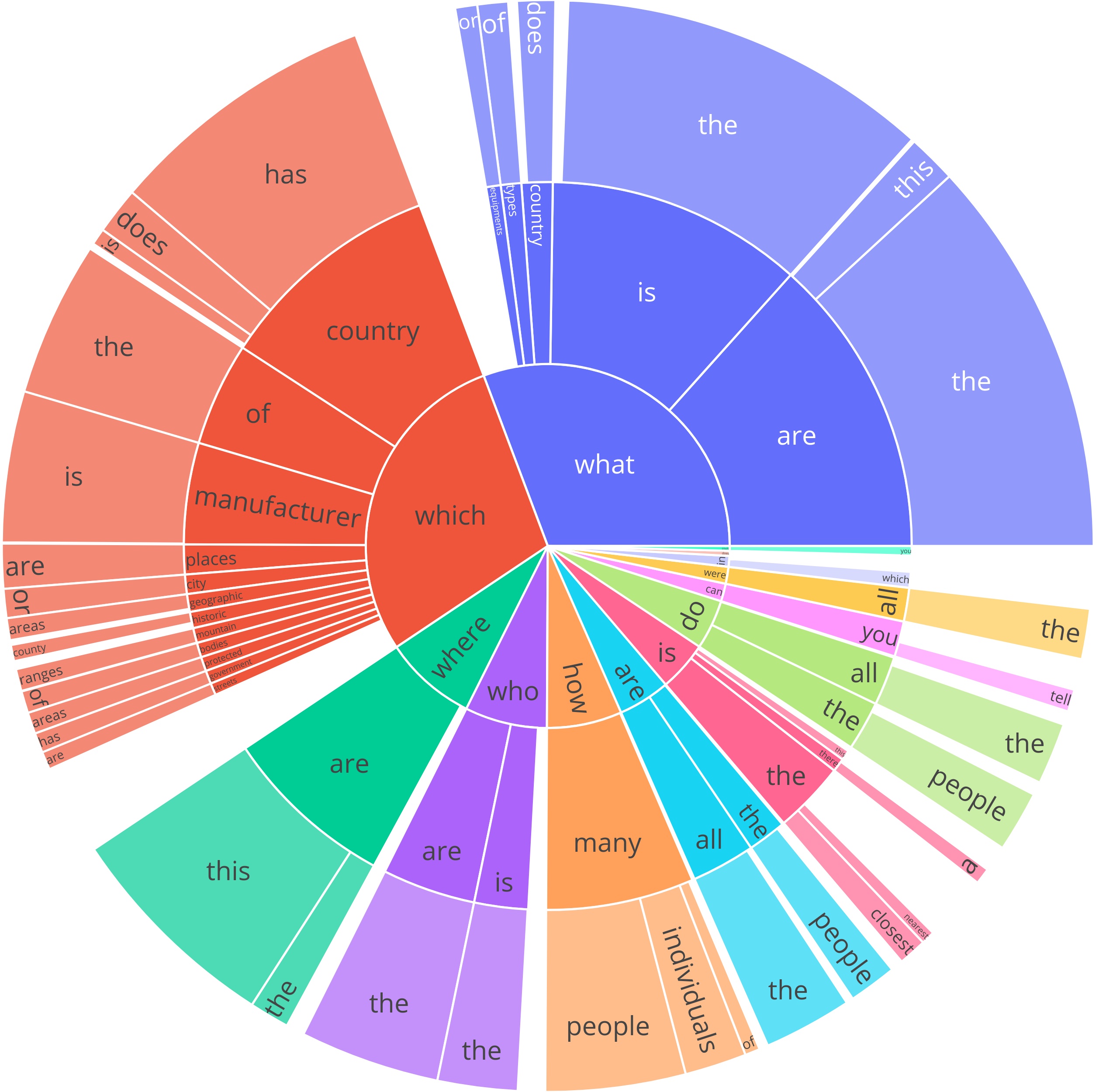}
  \caption{Types of questions covered in M$^3$-VQA.}
  \label{question_type}
\end{figure*}

We conducted heuristic identification for each question type in the dataset. To identify question types, we performed hierarchical parsing on each question. Each question sentence was split by spaces (interpreted as words or phrases), and up to the first three levels were used to build a hierarchical structure. We then counted the frequency of each path combination at each level and set a threshold of 50 to filter out low-frequency paths. As shown in Figure~\ref{question_type}, the dataset covers a diverse range of questions centered around locations, entities, events, comparisons, and numerical information.

\subsection{Data Type}
\begin{figure*}
  \centering
  \includegraphics[width=0.7\textwidth]{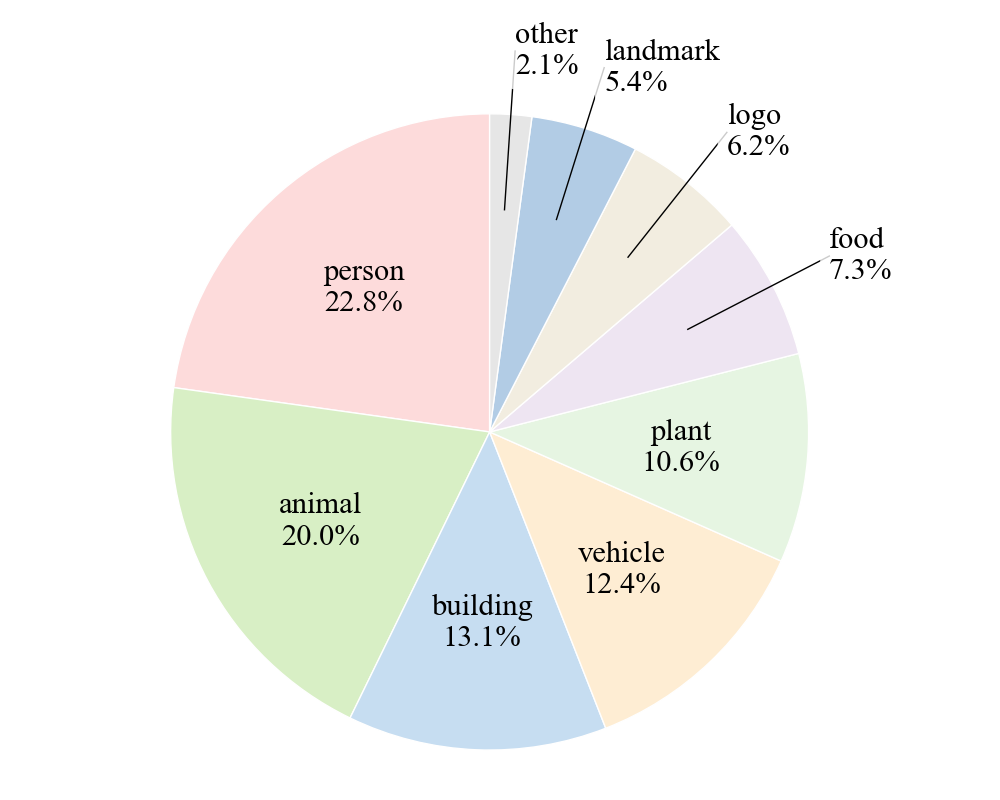}
  \caption{Types of data in M$^3$-VQA
  .}
  \label{data_type}
\end{figure*}

\begin{figure*}
  \centering
  \includegraphics[width=\textwidth]{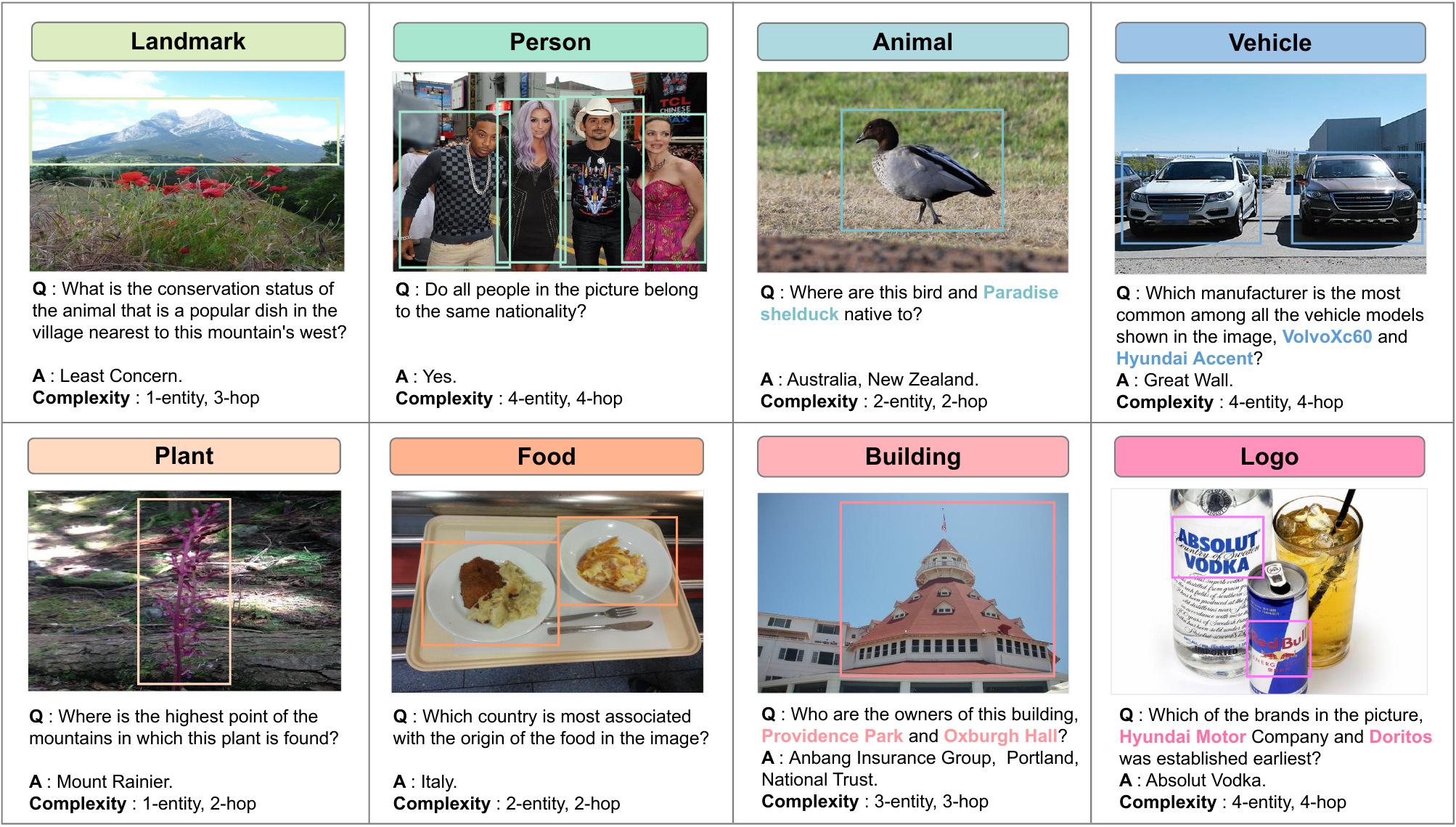}
  \caption{Examples of M$^3$-VQA questions across diverse fine-grained entity types.}
  \label{demo}
\end{figure*}

To gain a more comprehensive understanding of the dataset’s semantic composition, we conducted a categorical analysis of the subjects referenced by the questions associated with each image. The data were manually annotated and categorized into nine distinct semantic classes: person, animal, building, vehicle, plant, food, logo, landmark, and other. The distribution across these categories reflects a broad coverage of real-world entities, which enhances the model’s generalization capability in diverse scenarios. As shown in Figure~\ref{data_type} and Figure~\ref{demo}, the most frequent categories are person (2,406 instances), animal (2,115), and building (1,388), followed by vehicle (1,312) and plant (1,122). Less common yet still representative categories include food (772), logo (652), and landmark (574). The “other” category, consisting of samples that are difficult to classify or semantically ambiguous, contains 224 instances. This diverse category distribution supports comprehensive and robust model evaluation across a wide range of semantic domains.

\section{Experimental Details}

\subsection{Benchmarking Protocols}
We introduce six benchmarking protocols to assess model performance under varying levels of information accessibility. The table~\ref{setting_comparison}  and table~\ref{setting} compare these settings.

\begin{table*}

  \centering
  \begin{tabular}{lcccccccccc}
\toprule
\textbf{}
& \textbf{Sentence}
& \textbf{Section}
& \textbf{Entity-Name} 
& \textbf{KB} 
& \textbf{Original} 
& \textbf{Q-Only} \\
\midrule 

I & \ding{51}\ & \ding{51} & \ding{51} & \ding{51} & \ding{51} & \ding{55} \\

Q & \ding{51} & \ding{51} & \ding{51} & \ding{51} & \ding{51} & \ding{51} \\

Visual entity name & \ding{51} & \ding{51} & \ding{51} & Retrieved & \ding{55} & \ding{55} \\

Evidence & Sentence & Section & \ding{55} & Retrieved & \ding{55} & \ding{55} \\

\bottomrule
\end{tabular}
  \caption{Comparison of different benchmarking protocols.}
  \label{setting_comparison}
\end{table*}

\begin{table*}
  \centering
  \small
  \begin{tabular}{lcccccccccc}
\toprule
\textbf{Setting}
&\textbf{Input}
&\textbf{Methods} 
&\textbf{Example Models} 
&\textbf{Provided Evidence} 
&\textbf{Knowledge Base} \\
\midrule 

\textit{Original} & \{I, Q\} & End-to-end & Qwen2.5-VL-72B-Instruct & No & - \\

\textit{Oracle} & \{I, Q, E\} & End-to-end & Qwen2.5-VL-72B-Instruct & Yes & - \\

\textit{KB} &  \{I, Q, K\} & Pipeline & Heuristic Retrieval, Agentic Retrieval & No & Wikipedia \\

\bottomrule
\end{tabular}
\caption{Comparison of different evaluation settings.}
\label{setting}
\end{table*}

\subsection{Agentic Retrieval}
Heuristic retrieval methods suffer from overloaded queries. Using the full question as a single text query is problematic when the question contains multiple entities or sequential sub-questions. Similarly, using the entire image may hinder visual retrieval, especially when multiple entities are present. These one-shot strategies place excessive burden on a single query, often retrieving superficially relevant but unhelpful content.

To handle complex multi-entity, multi-hop questions in the KB setting, we implement Agentic Retrieval, a multi-agent retrieval-augmented generation framework. The core idea is to emulate how humans break down and solve complex problems in steps. The framework consists of three components: Planner, Executor, and Solver. The Executor includes tools such as object detection, text retrieval, and image retrieval. The Planner devises a strategy and calls the appropriate tools. The Solver generates the final answer based on the retrieved information.

\paragraph{Planner}
The Planner is the core module, responsible for decomposing the problem and planning the next steps based on feedback from the Executor.
It first calls the object detection module to locate relevant objects in the image, then segments the image accordingly. It may then perform image retrieval on the segments or issue single-hop queries for text retrieval. Once sufficient information is gathered, it passes the results to the Solver.

\paragraph{Executor}
The Executor is responsible for executing tool calls. It includes not only the text and image retrieval modules from heuristic retrieval, but also an object detection module to identify object coordinates in the image based on specific criteria.

\paragraph{Solver}
The Solver compiles the outputs from retrieval tools, removing tool execution traces and retaining only the retrieved content. It then attempts to answer the original question based on this information.

The entire process is fully automated and runs iteratively until the system determines that sufficient knowledge has been gathered to generate the final answer.

\subsection{Text Retrieval}
In both the heuristic retrieval and agentic retrieval models, we use text retrieval. We employ bge-large-en-v1.5 \cite{xiao2024c} to compute embeddings for the text, which is an advanced and commonly used text embedding model. For knowledge base chapters, we directly compute embeddings and index them using FAISS \cite{douze2024faiss}. For queries, we append instructions and compute embeddings, then use FAISS’s cosine similarity to retrieve the top 10 closest chapters. The instruction is:
"Represent this sentence for searching relevant passages:"

\subsection{Image Retrieval}
For image retrieval, we use clip-vit-large \cite{radford2021learning} to compute embeddings. We compute embeddings separately for the article summary and its image, then sum the two embeddings to form the entity embedding, which is indexed using FAISS \cite{douze2024faiss}. Based on our preliminary experiments, the combined embedding of both the summary and the image outperforms using either image or summary alone, as well as combining entity names and image embeddings. More effective embedding calculation methods will be explored in future research. For image queries, we directly compute the image embedding, then use FAISS’s cosine similarity to retrieve the top 10 closest entities. We then use the question as a query in the text retrieval module to search for the 10 most similar chapters from the 10 corresponding articles as retrieval results.

\subsection{Object Detection}
In the agentic retrieval model, we incorporate an object detection module. We chose Qwen2.5-VL-7B-Instruct \cite{bai2025qwen2}, which uses a variety of rectangular box and point-based techniques for general object localization, enabling hierarchical localization and standardized JSON format output. The prompt we use is:
"Outline the position of {query} and output all the bbox coordinates in JSON format."

\subsection{Backbone Models}
The backbone models we selected have the following characteristics: state-of-the-art; both open-source and closed-source; from different brands; with varying parameter sizes; multimodal and pure language models. The specific model introductions are as follows.

\paragraph{Qwen2.5-VL-3B, 7B, 72B-Instruct}Released in January 2025. This model \cite{bai2025qwen2} makes significant progress in world perception, acting as a visual agent, understanding long videos, capturing events, visual localization, and structured outputs.

\paragraph{Qwen2.5-VL-32B-Instruct}Released in March 2025. Based on the Qwen2.5-VL series, this model \cite{bai2025qwen2} has been optimized using reinforcement learning for responses that better align with human subjective preferences, enhancing mathematical reasoning, fine-grained image understanding, and reasoning capabilities.

\paragraph{Qwen2-VL}Released in August 2024. This model \cite{wang2024qwen2} is Capable of understanding images with various resolutions and aspect ratios, processing long videos over 20 minutes, and acting as a visual agent for mobile devices and robots, supporting multiple languages.

\paragraph{LLaVA-OneVision-7B}Released in August 2024. LLaVA-OneVision \cite{li2024llava} is an open-source multimodal LLM that trains Qwen2 on multimodal instruction-following data generated by GPT. It is the first model that can break the performance limits of open LMMs across three significant computing environments: single image, multi-image, and video scenarios. Notably, LLaVA-OneVision's design enables strong transfer learning across different modalities/scenarios, showcasing powerful video understanding and cross-scenario capabilities by transferring tasks from images to videos.

\paragraph{InternVL2.5}Released in December 2024. This \cite{chen2024expanding} is an advanced multimodal large language model (MLLM) series built on InternVL2\_0, significantly enhanced in training and testing strategies as well as data quality, while retaining its core "ViT-MLP-LLM" architecture.

\paragraph{MiniCPM-V-2.6}Released in August 2024. The most powerful model in the MiniCPM-V series, with 8B parameters, it \cite{yao2024minicpm} outperforms GPT-4V in single-image, multi-image, and video understanding, excelling in single-image understanding compared to GPT-4o mini, Gemini 1.5 Pro, and Claude 3.5 Sonnet, and is the first to support real-time video understanding on an iPad.

\paragraph{DeepSeek-VL2}Released in December 2024. This \cite{wu2024deepseek} is an advanced large expert mixture (MoE) visual language model with 4.5B active parameters, significantly improved over its predecessor, DeepSeek-VL. DeepSeek-VL2 demonstrates outstanding performance in various tasks, including visual question answering, optical character recognition, document/table/chart understanding, and visual grounding.

\paragraph{Llama-3.1-Instruct}Released in July 2024. Llama 3.1 \cite{grattafiori2024llama} is an autoregressive language model using an optimized transformer architecture. The adjusted version uses supervised fine-tuning (SFT) and reinforcement learning with human feedback (RLHF) to align with human preferences for utility and safety.

\paragraph{Qwen2.5-Instruct}Released in September 2024. Qwen2.5 \cite{yang2024qwen2} is the latest series in the Qwen large language model family, significantly improving encoding and mathematical abilities. It also shows marked improvement in instruction-following, generating long texts, understanding structured data, and generating structured outputs. The model is more resilient to system prompt diversity and enhances chatbot role-playing and condition setting.

\paragraph{GPT-4o}GPT-4o \cite{hurst2024gpt} is the fourth-generation multimodal large language model released by OpenAI in 2024, further optimized from GPT-4. It not only understands and generates natural language but also processes image, audio, and video inputs, enabling cross-modal information integration. GPT-4o features significant improvements in computational efficiency and inference speed, supporting real-time interaction. It is suitable for a wide range of applications, including conversational assistants, creative design, intelligent education, and visual search. The model is trained on massive datasets comprising text, images, and multimedia content, and leverages advanced self-supervised learning and reinforcement learning strategies. As a result, GPT-4o achieves industry-leading performance in accuracy, coherence, and safety.

\subsection{Image Descriptions}
Since pure language models cannot directly process images, we use image descriptions as a substitute input for the pure language models. Specifically, we use Qwen2.5-VL-7B-Instruct \cite{bai2025qwen2} to generate detailed descriptions of the images.

\section{Evaluation Metrics}

Following previous work \cite{chen2023can}, we use answer aliases from Wikidata as multiple reference answers for string-based questions. Following previous work \cite{goyal2017making, marino2019ok, chen2023can}, we allow a one-year error margin for temporal questions, as historical events often have approximate dates. Since some questions may have multiple valid answers, we use Intersection over Union (IoU) accuracy, following previous work \cite{mensink2023encyclopedic}. This is more accurate than exact match, as exact match overlooks partial correctness and cannot differentiate between completely wrong and partially correct answers. The arithmetic mean of all question accuracies is taken as the final score.

\subsection{Maximum Matching}
Since we provide aliases for some answers, we cannot directly compute the IoU between two sets. Our evaluation metric is essentially a maximum bipartite matching problem. For example, suppose the predicted answers are $[a1, b4, a3, d1]$, and the candidate answers are $[[a1, a2, a3], [b1, b2], [c1]]$. Each sublist in the candidate answers represents an answer and its aliases. We treat the predicted answers as node set $X$, with each item as a node in $X$. The candidate answers form node set $Y$, where each sublist is a node in $Y$. If an item in the predicted answers belongs to a sublist in $Y$, we draw an edge between the corresponding nodes. This forms a bipartite graph. The size of the maximum matching in this graph is denoted as $NUM$. Then, the IoU accuracy can be calculated as:
$accuracy = NUM / (|X| + |Y| - NUM)$.
In our example, $NUM = 1$ (note that both $a1$ and $a3$ connect to $[a1, a2, a3]$, but only one match is allowed). So, $accuracy = 1 / (4 + 3 - 1)$.
The optimal algorithm for maximum matching is the Hopcroft-Karp algorithm, with a time complexity of $O(\sqrt{V}, E)$.

\subsection{Efficient Evaluation}
Since some answers have many aliases, calculating IoU accuracy can be time-consuming for certain questions. In our practice, it is rare for two predicted answers to belong to the same alias group in the candidate answers (e.g., both $a1$ and $a3$ appearing simultaneously). Therefore, we provide an alternative method for calculating IoU accuracy. Specifically, we flatten the candidate answers into a one-dimensional list and count how many predicted items exist in this flattened list, denoted as NUM’. Then, IoU accuracy is:
$accuracy = NUM’ / (|X| + |Y| - NUM’)$.
In our tests, this method differs from the maximum matching-based approach by less than 0.2\%, while significantly reducing evaluation time — completing evaluation on 13K items in just a few seconds.

\subsection{Other Common Evaluation Metrics}
INFOSEEK \cite{chen2023can} uses Wikidata to obtain multiple reference answers. For temporal and numerical questions, it adopts lenient accuracy and computes the harmonic mean across dataset splits as the overall accuracy. We use a similar method but extend it to support multiple answers.
Dyn-VQA \cite{li2024benchmarking} tokenizes both generated and gold answers into token lists, then calculates the ratio of generated tokens found in the gold token list as the accuracy. However, this method has a drawback: longer or more verbose answers may receive higher scores even if additional content is incorrect. In contrast, the IoU metric penalizes extra incorrect answers — guessing more answers may lead to a lower score.
EVQA \cite{mensink2023encyclopedic} employs BERT-based matching for evaluation, which is computationally expensive, especially for our larger dataset (13K items) and numerous candidate answers.
Some works also use GPT-based evaluation, but due to the number of candidate answers in our setting, this approach is costly and less stable.

\subsection{Quality of the IoU Metric}
To verify that our IoU metric reliably reflects the model's actual performance, we introduce human-based accuracy. Specifically, we sample 200 predicted answers from our experiments and ask an expert to evaluate their correctness. We compare these expert evaluations with the scores given by the IoU metric. We find that in 95\% of cases, the IoU-based score agrees with human judgment. This demonstrates the effectiveness of the IoU evaluation metric. At the same time, as an automated metric, it offers significantly lower computational cost and better scalability.

\section{Additional Experiment}
\label{appx:H}

\subsection{Additional Experimental Results}

\begin{table*}
  \centering
  \small
  \begin{tabular}{clccccccccc}
\toprule
\multirow{2}{*}{\textbf{Setting}} & \multirow{2}{*}{\textbf{Model}} 
& \multicolumn{4}{c}{\textbf{Hop}} 
& \multicolumn{4}{c}{\textbf{Entity}} 
& \multirow{2}{*}{\textbf{All}} \\
\cmidrule(lr){3-6}
\cmidrule(lr){7-10}
& & \textbf{1} & \textbf{2} & \textbf{3} & \textbf{4+} 
  & \textbf{1} & \textbf{2} & \textbf{3} & \textbf{4+} 
  &  \\
\midrule

\multirow{18}{*}{Sentence}
& Qwen2.5-VL-3B-Instruct        & 71.30 & 48.47 & 41.11 & 32.55 & 55.99 & 48.25 & 40.10 & 32.55 & 42.60  \\
& Qwen2.5-VL-7B-Instruct        & 67.77 & 54.52 & 49.52 & 39.52 & 60.51 & 54.98 & 47.06 & 39.52 & 48.97  \\
& Qwen2.5-VL-32B-Instruct       & 73.42 & 57.58 & 56.66 & 42.03 & 63.09 & 59.15 & 55.72 & 42.03 & 53.48  \\
& Qwen2.5-VL-72B-Instruct       & 71.78 & 64.18 & 63.86 & 45.01 & 63.46 & 66.75 & 65.10 & 45.01 & 58.41  \\
& Qwen2-VL-7B-Instruct          & 57.70 & 46.64 & 39.75 & 34.58 & 51.66 & 47.13 & 36.92 & 34.58 & 41.21  \\
& InternVL2.5-1B               & 31.50 & 22.06 & 19.89 & 17.01 & 21.40 & 24.59 & 21.22 & 17.01 & 20.44  \\ 

& InternVL2.5-2B               & 49.77 & 37.55 & 32.87 & 30.02 & 43.73 & 36.92 & 30.66 & 30.02 & 34.47  \\
& InternVL2.5-4B               & 67.88 & 47.34 & 40.86 & 35.04 & 51.40 & 50.05 & 40.46 & 35.04 & 42.77  \\
& InternVL2.5-8B               & 60.63 & 52.24 & 46.89 & 39.26 & 51.36 & 54.18 & 47.62 & 39.26 & 46.82  \\
& InternVL2.5-26B               & 58.31 & 54.92 & 48.44 & 41.32 & 52.06 & 56.01 & 49.56 & 41.32 & 48.50  \\
& InternVL2.5-38B               & 77.24 & 64.78 & 62.85 & 44.56 & 64.16 & 67.15 & 65.18 & 44.56 & 58.51  \\
& InternVL2.5-78B	             & 73.73 & 63.67 & 60.71 & 49.12 & 56.73 & 67.40 & 66.46 & 49.12 & 58.71   \\
& LLaVA-OneVision-7B & 71.43 & 49.76 & 47.10 & 39.13 & 54.47 & 51.18 & 48.20 & 39.13 & 47.15  \\
& DeepSeek-VL2                & 66.31 & 47.34 & 43.58  & 38.27 & 51.08 & 50.56 & 43.33 & 38.27 & 44.64  \\
& MiniCPM-V-2.6                & 64.30 & 51.45 & 46.34 & 39.77 & 55.27 & 53.67 & 44.47 & 39.77 & 46.92  \\
& GPT-4o & 73.86 & 61.40 & 63.06 & 48.04 & 59.63 & 64.23 & 66.97 & 48.04 & 58.63 \\
& Llama-3.1-8B-Instruct         & 61.05 & 46.53 & 40.57 & 36.13 & 53.79 & 45.58 & 38.05 & 36.13 & 42.26  \\
& Qwen2.5-7B-Instruct           & 62.02 & 47.62 & 44.60 & 34.67 & 55.31 & 47.45 & 41.94 & 34.67 & 43.52  \\
\midrule

\multirow{18}{*}{Section}
& Qwen2.5-VL-3B-Instruct        & 69.08 & 43.23 & 38.40 & 30.99 & 50.82 & 44.64 & 37.87 & 30.99 & 39.69 \\
& Qwen2.5-VL-7B-Instruct        & 53.55 & 47.36 & 44.65 & 34.32 & 50.15 & 49.51 & 42.26 & 34.32 & 42.66  \\
& Qwen2.5-VL-32B-Instruct       & 70.86 & 52.17 & 53.69 & 39.28 & 56.83 & 55.87 & 53.78 & 39.28 & 50.02  \\
& Qwen2.5-VL-72B-Instruct       & 67.03 & 56.45 & 56.44 & 41.27 & 53.79 & 60.97 & 59.10 & 41.27 & 52.35  \\
& Qwen2-VL-7B-Instruct          & 55.86 & 43.08 & 37.75 & 32.77 & 47.68 & 45.00 & 35.35 & 32.77 & 38.92  \\
& InternVL2.5-1B               & 31.57 & 21.48 & 18.24 & 12.28 & 20.91 & 24.49 & 19.13 & 12.28 & 18.18   \\
& InternVL2.5-2B               & 48.62 & 36.24 & 29.32 & 27.31 & 37.37 & 36.99 & 29.93 & 27.31 & 31.94  \\
& InternVL2.5-4B               & 66.01 & 41.32 & 37.66 & 32.91 & 47.17 & 44.31 & 37.53 & 32.91 & 39.34  \\
& InternVL2.5-8B               & 58.61 & 46.03 & 42.30 & 34.85 & 47.23 & 48.47 & 42.71 & 34.85 & 42.12  \\
& InternVL2.5-26B               & 55.25 & 49.61 & 46.21 & 39.23 & 46.42 & 52.13 & 48.18 & 39.23 & 45.50  \\
& InternVL2.5-38B               & 73.21 & 59.30 & 57.28 & 41.60 & 56.66 & 62.96 & 60.92 & 41.60 & 53.95  \\
& InternVL2.5-78B	             & 72.12 & 59.84 & 56.93 & 45.54 & 55.03 & 63.92 & 61.39 & 45.54 & 55.16    \\
& LLaVA-OneVision-7B & 69.87 & 46.10 & 42.82 & 35.96 & 50.15 & 47.93 & 44.73 & 35.96 & 43.62  \\
& DeepSeek-VL2                & 63.04 & 41.31 & 38.49 & 31.17 & 47.09 & 44.88 & 37.40 & 31.17 & 38.81   \\
& MiniCPM-V-2.6                & 60.77 & 39.93 & 40.14 & 31.91 & 47.75 & 40.85 & 39.48 & 31.91 & 39.11  \\
& GPT-4o & 71.03 & 54.57 & 56.07 & 44.22  & 53.99 & 59.22 & 59.07 & 44.22 & 53.07 \\
& Llama-3.1-8B-Instruct         & 58.36 & 41.87 & 39.65 & 35.82 & 48.60 & 41.86 & 38.53 & 35.82 & 40.49  \\
& Qwen2.5-7B-Instruct           & 59.38 & 41.90 & 39.02 & 31.26 & 51.16 & 41.68 & 36.09 & 31.26 & 38.87  \\

\midrule
\multirow{18}{*}{Name}
& Qwen2.5-VL-3B-Instruct         & 49.40 & 25.91 & 22.98 & 20.20 & 29.57 & 30.30 & 23.67 & 20.20 & 24.98 \\
& Qwen2.5-VL-7B-Instruct         & 41.26 & 35.62 & 34.80 & 23.49 & 28.35 & 43.13 & 37.77 & 23.49 & 31.83  \\
& Qwen2.5-VL-32B-Instruct        & 62.99 & 38.25 & 42.18 & 34.36 & 38.11 & 46.50 & 45.73 & 34.36 & 40.39  \\
& Qwen2.5-VL-72B-Instruct        & 65.20 & 42.52 & 48.37 & 38.29 & 39.55 & 52.82 & 52.90 & 38.29 & 45.05  \\
& Qwen2-VL-7B-Instruct           & 40.33 & 28.85 & 29.99 & 28.23 & 26.50 & 34.82 & 32.08 & 28.23 & 30.00  \\
& InternVL2.5-1B               & 36.08 & 20.28 & 22.02 & 13.90 & 18.76 & 26.35 & 24.78 & 13.90 & 20.10  \\ 
& InternVL2.5-2B               & 34.48 & 19.54 & 25.21 & 20.92 & 20.95 & 24.86 & 26.75 & 20.92 & 23.20  \\ 
& InternVL2.5-4B               & 51.81 & 27.14 & 29.90 & 25.34 & 28.83 & 33.46 & 32.81 & 25.34 & 29.56  \\ 
& InternVL2.5-8B                & 41.50 & 26.53 & 25.39 & 21.44 & 25.00 & 32.55 & 27.24 & 21.44 & 25.71  \\
& InternVL2.5-26B               & 46.45 & 34.75 & 35.51 & 28.97 & 28.43 & 43.00 & 39.45 & 28.97 & 34.09  \\ 
& InternVL2.5-38B               & 65.26 & 43.15 & 46.81 & 37.33 & 39.54 & 54.07 & 50.77 & 37.33 & 44.35  \\ 
& InternVL2.5-78B	             & 63.26 & 44.21 & 47.04 & 38.87 & 38.92 & 55.15 & 51.23 & 38.87 & 45.02   \\
& LLaVA-OneVision-7B& 49.68 & 30.79 & 33.21 & 26.20 & 28.23 & 38.93 & 36.68 & 26.20 & 31.69  \\
& DeepSeek-VL2                & 50.67 & 29.96 & 31.45 & 24.97 & 28.79 & 37.24 & 34.62 & 24.97 & 30.56  \\ 
& MiniCPM-V-2.6                 & 53.61 & 32.28 & 32.66 & 30.49 & 30.85 & 39.52 & 35.99 & 30.49 & 33.60  \\
& GPT-4o & 69.69 & 48.18 & 54.59 & 45.29 & 42.02 & 60.65 & 60.08 & 45.29 & 51.21 \\
& Llama-3.1-8B-Instruct          & 51.01 & 28.81 & 29.67 & 25.09 & 30.31 & 35.10 & 31.51 & 25.10 & 29.73  \\
& Qwen2.5-7B-Instruct            & 45.45 & 26.70 & 28.77 & 20.59 & 28.20 & 33.28 & 29.72 & 20.59 & 26.97  \\


\bottomrule
\end{tabular}
  \caption{Performance comparison of various models under different settings.}
  \label{model_performance_additional}
\end{table*}

\begin{table*}
  \centering
  \small
  \begin{tabular}{clccccccccc}
\toprule
\multirow{2}{*}{\textbf{Setting}} & \multirow{2}{*}{\textbf{Model}} 
& \multicolumn{4}{c}{\textbf{Hop}} 
& \multicolumn{4}{c}{\textbf{Entity}} 
& \multirow{2}{*}{\textbf{All}}  \\
\cmidrule(lr){3-6}
\cmidrule(lr){7-10}
& & \textbf{1} & \textbf{2} & \textbf{3} & \textbf{4+} 
  & \textbf{1} & \textbf{2} & \textbf{3} & \textbf{4+} 
  &  \\
\midrule

\multirow{36}{*}{KB}
& \textit{Heuristic Retrieval} \\
& Qwen2.5-VL-3B-Instruct         & 38.47 & 18.13 & 21.82 & 25.37 & 23.84 & 23.63 & 22.66 & 25.37 & 23.48  \\
& Qwen2.5-VL-7B-Instruct         & 41.62 & 22.33 & 23.38 & 26.54 & 26.07 & 24.40 & 25.11 & 26.54 & 25.68  \\
& Qwen2.5-VL-32B-Instruct        & 48.49 & 29.41 & 32.83 & 31.13 & 31.20 & 34.60 & 34.83 & 31.13 & 32.75  \\
& Qwen2.5-VL-72B-Instruct        & 51.65 & 33.67 & 36.49 & 34.99 & 33.51 & 39.69 & 39.08 & 34.99 & 36.57  \\
& Qwen2-VL-7B-Instruct           & 31.27 & 19.21 & 21.21 & 25.14 & 20.96 & 21.64 & 22.42 & 25.14 & 22.85  \\
& InternVL2.5-1B               & 27.03 & 9.49 & 13.49 & 12.47 & 13.71 & 10.10 & 15.90 & 12.47 & 13.31  \\ 
& InternVL2.5-2B               & 21.69 & 14.78 & 18.21 & 20.64 & 16.65 & 16.12 & 18.81 & 20.64 & 18.47  \\ 
& InternVL2.5-4B               & 32.56 & 17.70 & 20.58 & 23.55 & 21.80 & 19.54 & 21.27 & 23.55 & 21.85  \\ 
& InternVL2.5-8B                & 30.46 & 20.33 & 24.61 & 28.49 & 22.28 & 22.87 & 25.87 & 28.49 & 25.33  \\
& InternVL2.5-26B               & 31.81 & 24.36 & 29.00 & 34.45 & 25.23 & 26.91 & 30.15 & 34.45 & 29.90  \\
& InternVL2.5-38B               & 45.34 & 28.50 & 32.70 & 32.40 & 31.42 & 31.79 & 34.40 & 32.40 & 32.63  \\ 
& InternVL2.5-78B	             & 44.70 & 28.82 & 33.80 & 32.77 & 29.82 & 33.32 & 36.27 & 32.77 & 33.16   \\
& LLaVA-OneVision-7B& 37.62 & 19.75 & 24.85 & 28.37 & 26.80 & 27.03 & 24.91 & 28.37 & 25.83  \\
& MiniCPM-V-2.6                 & 35.72 & 19.47 & 24.98 & 25.44 & 24.99 & 23.88 & 25.84 & 25.44 & 24.69  \\
& GPT-4o                         & 54.93 & 30.31 & 33.17 & 33.80 & 35.28 & 34.52 & 34.64 & 33.80 & 34.49   \\
& Llama-3.1-8B-Instruct          & 30.73 & 18.91 & 20.13 & 28.23 & 19.10 & 23.56 & 22.22 & 28.23 & 23.37  \\
& Qwen2.5-7B-Instruct            & 33.32 & 18.66 & 21.53 & 26.26 & 22.17 & 23.92 & 22.41 & 26.26 & 23.37  \\
\cmidrule(lr){2-11}

& \textit{Agentic Retrieval} \\
& Qwen2.5-VL-3B-Instruct         & 42.72  & 20.86  & 24.25  & 27.25  & 26.67 & 22.66  & 25.80 & 27.25 & 25.95 \\
& Qwen2.5-VL-7B-Instruct         & 41.44 & 28.25 & 30.69 & 31.45 & 27.52 & 32.13 & 33.52 & 31.45 & 31.24   \\
& Qwen2.5-VL-32B-Instruct        & 52.72 & 34.50 & 36.49 & 36.12  & 33.98 & 40.42 & 39.32 & 36.12 & 37.24 \\
& Qwen2.5-VL-72B-Instruct        & 54.49 & 36.89 & 39.83 & 35.48 & 35.68 & 43.07 & 43.11 & 35.48 & 38.91  \\
& Qwen2-VL-7B-Instruct           & 36.94 & 23.34 & 25.96 & 27.37 & 23.68 & 27.19 & 28.11 & 27.37 & 26.70 \\
& InternVL2.5-1B                 & 32.92  & 11.37  & 15.39  & 12.17  & 15.78 & 12.76  & 18.40  & 12.17  & 14.82  \\
& InternVL2.5-2B                 & 28.34  & 17.31  & 19.29  & 21.47  & 18.67 & 19.74  & 20.52  & 21.47  & 20.28  \\
& InternVL2.5-4B                 & 37.24  & 18.03  & 20.63  & 23.84  & 21.52 & 21.01  & 22.40  & 23.84  & 22.44  \\
& InternVL2.5-8B                 & 39.53 & 26.15 & 27.86 & 30.97 & 28.32 & 28.45 & 29.13 & 30.97 & 29.44 \\
& InternVL2.5-26B                & 36.72 & 26.72 & 31.85 & 34.46  & 26.50 & 30.88 & 33.81 & 34.46 & 31.87 \\
& InternVL2.5-38B                & 49.83 & 31.87 & 36.15 & 33.67  & 33.04 & 36.91 & 38.61 & 33.67 & 35.44 \\
& InternVL2.5-78B	             & 19.97 & 32.52 & 36.90 & 35.20 & 31.38 & 39.04 & 40.25 & 35.20 & 36.37   \\
& LLaVA-OneVision-7B             & 42.73  & 24.07 & 28.35 & 29.38 & 28.28 & 27.32 & 29.61 & 29.38 & 28.85    \\
& MiniCPM-V-2.6                  & 42.84 & 23.08 & 27.90 & 28.02 & 27.11 & 25.92 & 30.00 & 28.02 & 28.01  \\
& GPT-4o                         & 55.36 & 35.73 & 39.66 & 36.06 & 35.34 & 42.77 & 42.61 & 36.06 & 38.83 \\
& Llama-3.1-8B-Instruct          & 38.17 & 23.93 & 25.92 & 31.43 & 24.78 & 26.74 & 28.29 & 31.43 & 28.26 \\
& Qwen2.5-7B-Instruct            & 36.26 & 20.10 & 22.11 & 27.45 & 22.55 & 22.59 & 23.95 & 27.45 & 24.55 \\

\bottomrule
\end{tabular}
  \caption{Performance comparison of various models under different settings.}
  \label{model_performance_additional_2}
\end{table*}

\begin{table*}
  \centering
  \small
  \begin{tabular}{clccccccccc}
\toprule
\multirow{2}{*}{\textbf{Setting}} & \multirow{2}{*}{\textbf{Model}} 
& \multicolumn{4}{c}{\textbf{Hop}} 
& \multicolumn{4}{c}{\textbf{Entity}} 
& \multirow{2}{*}{\textbf{All}}  \\
\cmidrule(lr){3-6}
\cmidrule(lr){7-10}
& & \textbf{1} & \textbf{2} & \textbf{3} & \textbf{4+} 
  & \textbf{1} & \textbf{2} & \textbf{3} & \textbf{4+} 
  &  \\
\midrule

\multirow{18}{*}{Original}
& Qwen2.5-VL-3B-Instruct         & 36.95 & 17.44 & 17.62 & 16.01 & 24.53 & 18.55 & 17.04 & 16.01 & 18.66  \\
& Qwen2.5-VL-7B-Instruct         & 34.87 & 20.66 & 21.63 & 21.74 & 24.14 & 24.30 & 21.08 & 21.74 & 22.53  \\
& Qwen2.5-VL-32B-Instruct        & 46.84 & 24.94 & 29.84 & 25.55 & 30.12 & 29.44 & 30.67 & 25.55 & 28.66  \\
& Qwen2.5-VL-72B-Instruct        & 47.94 & 29.48 & 34.19 & 29.20 & 30.81 & 35.51 & 36.16 & 29.20 & 32.55  \\
& Qwen2-VL-7B-Instruct           & 28.68 & 17.55 & 20.45 & 24.34 & 20.49 & 20.87 & 20.11 & 24.34 & 21.71  \\
& InternVL2.5-1B               & 18.09 & 2.63 & 4.31 & 2.76 & 7.20 & 3.60 & 5.02 & 2.76 & 4.54  \\
& InternVL2.5-2B               & 17.02 & 5.04 & 8.71 & 8.41 & 9.88 & 6.53 & 8.33 & 8.41 & 8.41  \\
& InternVL2.5-4B               & 35.01 & 13.25 & 17.97 & 19.73 & 18.37 & 16.96 & 19.23 & 19.73 & 18.82  \\
& InternVL2.5-8B                & 26.14 & 16.86 & 19.44 & 17.36 & 18.45 & 19.60 & 19.91 & 17.36 & 18.69  \\
& InternVL2.5-26B               & 27.40 & 13.76 & 15.24 & 17.01 & 17.92 & 15.42 & 15.31 & 17.01 & 16.48  \\
& InternVL2.5-38B               & 42.20 & 24.43 & 30.55 & 26.02 & 28.69 & 28.87 & 31.23 & 26.02 & 28.55  \\
& InternVL2.5-78B	             & 43.45 & 27.15 & 33.35 & 29.05 & 29.76 & 32.54 & 34.41 & 29.05 & 31.28   \\
& LLaVA-OneVision-7B & 34.98 & 17.53 & 21.56 & 22.54 & 21.26 & 20.87 & 22.72 & 22.54 & 22.02  \\
& DeepSeek-VL2                & 40.20 & 20.31 & 24.86 & 22.68 & 24.62 & 23.96 & 26.21 & 22.68 & 24.32  \\
& MiniCPM-V-2.6                 & 31.46 & 19.13 & 23.05 & 25.05 & 19.17 & 23.05 & 25.31 & 25.05 & 23.45  \\
& GPT-4o & 51.60 & 23.30 & 24.25 & 27.93 & 34.40 & 25.87 & 22.31 & 27.93 & 27.50 \\
& Llama-3.1-8B-Instruct          & 26.16 & 15.94 & 21.66 & 21.54 & 18.62 & 19.27 & 21.95 & 21.54 & 20.61  \\
& Qwen2.5-7B-Instruct            & 23.99 & 12.85 & 18.69 & 17.35 & 16.70 & 16.09 & 18.35 & 17.35 & 17.27  \\

\midrule
\multirow{18}{*}{Q-Only}
& Qwen2.5-VL-3B-Instruct         & 8.70 & 11.22 & 15.15 & 16.84 & 8.11 & 12.59 & 17.12 & 16.84 & 14.22  \\
& Qwen2.5-VL-7B-Instruct         & 4.99 & 9.45 & 14.00 & 20.84 & 4.56 & 12.19 & 16.16 & 20.84 & 14.39  \\
& Qwen2.5-VL-32B-Instruct        & 9.89 & 9.36 & 17.73 & 20.99 & 8.44 & 11.17 & 19.64 & 20.99 & 16.12  \\
& Qwen2.5-VL-72B-Instruct        & 9.16 & 10.23 & 14.38 & 23.90 & 5.92 & 13.69 & 16.54 & 23.90 & 16.06  \\
& Qwen2-VL-7B-Instruct           & 9.80 & 10.74 & 17.07 & 20.23 & 12.95 & 12.54 & 15.54 & 20.23 & 15.99  \\
& InternVL2.5-1B               & 7.76 & 7.34 & 11.23 & 8.36 & 4.93 & 9.83 & 12.82 & 8.36 & 9.06  \\ 
& InternVL2.5-2B               & 6.27 & 6.18 & 12.18 & 11.49 & 4.19 & 7.98 & 14.26 & 11.49 & 10.01  \\
& InternVL2.5-4B               & 2.66 & 6.76 & 11.42 & 14.96 & 2.73 & 8.62 & 13.53 & 14.96 & 10.72  \\ 
& InternVL2.5-8B                & 2.75 & 8.08 & 12.33 & 16.36 & 2.48 & 10.86 & 14.70 & 16.36 & 11.82  \\
& InternVL2.5-26B               & 4.17 & 9.92 & 13.80 & 21.14 & 4.32 & 12.72 & 15.97 & 21.14 & 14.47  \\ 
& InternVL2.5-38B               & 1.01 & 7.23 & 10.44 & 18.06 & 1.08 & 9.85 & 12.86 & 18.06 & 11.38  \\ 
& InternVL2.5-78B	             & 1.01 & 9.39 & 14.74 & 22.14 & 4.69 & 11.94 & 15.89	& 22.14 & 14.73   \\
& LLaVA-OneVision-7B            & 10.58 & 11.41 & 19.81 & 20.77 & 10.48 & 14.32 & 20.66 & 20.77 & 17.32  \\
& DeepSeek-VL2                & 9.57 & 9.99 & 14.23 & 17.90 & 6.91 & 12.59 & 16.13 & 17.90 & 14.02  \\ 
& MiniCPM-V-2.6                 & 10.71 & 14.02 & 19.55 & 22.72 & 13.69 & 16.24 & 18.85 & 22.72 & 18.51  \\
& GPT-4o & 4.67 & 13.16 & 19.11 & 28.76 & 7.35 & 17.39 & 20.28 & 28.76 & 19.63  \\
& Llama-3.1-8B-Instruct          & 9.89 & 9.36 & 17.73 & 20.99 & 8.44 & 11.17 & 19.64 & 20.99 & 16.12  \\
& Qwen2.5-7B-Instruct            & 2.83 & 7.70 & 12.51 & 17.44 & 3.86 & 10.40 & 13.79 & 17.44 & 12.16 \\

\bottomrule
\end{tabular}
  \caption{Performance comparison of various models under different settings.}
  \label{model_performance_additional_3}
\end{table*}

We benchmark a wide range of models. This includes advanced closed-source models like GPT-4o \cite{hurst2024gpt}, and open-source models such as Qwen2.5-VL-72B-Instruct \cite{bai2025qwen2}. To compare visual models from different providers, we evaluate 7B/8B versions of Qwen2.5-VL, Qwen2-VL \cite{wang2024qwen2}, InternVL2.5 \cite{chen2024expanding}, Llava-OneVision \cite{li2024llava}, DeepSeek-VL2 \cite{wu2024deepseek} and MiniCPM-V-2.6 \cite{yao2024minicpm}. To study the impact of model size, we test the 3B, 7B, 32B, and 72B versions of Qwen2.5-VL, as well as the 1B, 2B, 4B, 8B, 26B, 38B, and 78B versions of InternVL2.5.  We also evaluate pure language models (e.g., LLaMA-3.1 \cite{grattafiori2024llama}, Qwen2.5 \cite{yang2024qwen2}) by replacing image inputs with textual descriptions. Due to the high computational cost, each experiment was conducted only once.

We provide the performance of additional models in Tables~\ref{model_performance_additional}, \ref{model_performance_additional_2} and \ref{model_performance_additional_3}. Please note that due to the context length limitations of certain models, results under the KB setting may be missing, and their performance under the Section setting may also be affected (e.g., DeepSeek-VL2).

\subsection{Attribution Analysis of Retrieval Models}
\begin{table*}
  \centering
  \begin{tabular}{llcccccc}
\toprule
\multirow{2}{*}{\textbf{Hop}} 
& \multirow{2}{*}{\textbf{Top-k}} 
& \multicolumn{2}{c}{\textbf{Text}} 
& \multicolumn{2}{c}{\textbf{Image}} 
& \multicolumn{2}{c}{\textbf{All}} \\
\cmidrule(lr){3-4}
\cmidrule(lr){5-6}
\cmidrule(lr){7-8}
& & \textbf{Page} & \textbf{Section} 
& \textbf{Page} & \textbf{Section}
& \textbf{Page} & \textbf{Section} \\
\midrule

\multirow{4}{*}{all} 
& 3  & 23.11 & 20.25 & 19.41 & 9.93 & 41.88 & 28.96 \\
& 5  & 26.36 & 23.76 & 23.49 & 13.97 & 48.79 & 36.23 \\
& 7  & 28.04 & 25.56 & 26.45 & 17.37 & 52.98 & 41.20 \\
& 10 & \textbf{29.85} & \textbf{27.64} & \textbf{29.56} & \textbf{21.27} & \textbf{57.21} & \textbf{46.92} \\

\midrule
\multirow{4}{*}{1 hop} 
& 3 & 0.92 & 0.64 & 37.64 & 25.00 & 38.00 & 25.09 \\
& 5 & 1.28 & 0.73 & 46.25 & 32.88 & 46.79 & 33.06 \\
& 7 & 2.01 & 1.47 & 50.92 & 37.73 & 51.65 & 38.28 \\
& 10 & \textbf{2.38} & \textbf{1.74} & \textbf{55.22} & \textbf{44.41} & \textbf{56.95} & \textbf{45.05} \\

\midrule
\multirow{4}{*}{2 hop} 
& 3 & 20.47 & 19.10 & 26.94 & 13.94 & 46.15 & 30.97 \\
& 5 & 22.12 & 20.98 & 32.33 & 19.05 & 52.42 & 37.35 \\
& 7 & 23.20 & 21.95 & 36.22 & 23.64 & 56.63 & 42.49 \\
& 10 & \textbf{24.36} & \textbf{23.10} & \textbf{40.29} & \textbf{29.02} & \textbf{60.69} & \textbf{48.48} \\

\midrule
\multirow{4}{*}{3 hop} 
& 3 & 18.99 & 16.08 & 18.07 & 6.72 & 36.52 & 21.96 \\
& 5 & 20.30 & 18.67 & 21.46 & 10.32 & 41.58 & 27.97 \\
& 7 & 21.69 & 19.44 & 24.08 & 13.50 & 44.70 & 31.80 \\
& 10 & \textbf{23.08} & \textbf{21.12} & \textbf{26.88} & \textbf{17.34} & \textbf{48.57} & \textbf{37.23} \\

\midrule
\multirow{4}{*}{4 hop} 
& 3 & 35.06 & 30.47 & 10.64 & 6.52 & 45.37 & 35.85 \\
& 5 & 41.63 & 37.02 & 13.31 & 9.30 & 54.24 & 44.94 \\
& 7 & 44.93 & 40.79 & 15.54 & 11.65 & 59.37 & 50.91 \\
& 10 & \textbf{48.02} & \textbf{44.46} & \textbf{17.96} & \textbf{13.82} & \textbf{64.09} & \textbf{56.47} \\

\bottomrule
\end{tabular}
  \caption{Retrieval accuracy of the heuristic retrieval model under different hop and top-k settings.}
  \label{retrieval_performance_hops}
\end{table*}

\begin{table*}
  \centering
  \begin{tabular}{lcccccc}
\toprule
\multirow{2}{*}{\textbf{Hop}} 
& \multicolumn{2}{c}{\textbf{Text}} 
& \multicolumn{2}{c}{\textbf{Image}} 
& \multicolumn{2}{c}{\textbf{All}} \\
\cmidrule(lr){2-3}
\cmidrule(lr){4-5}
\cmidrule(lr){6-7}
& \textbf{Page} & \textbf{Section} 
& \textbf{Page} & \textbf{Section}
& \textbf{Page} & \textbf{Section} \\
\midrule

all   & \textbf{30.30} & \textbf{28.73} & \textbf{35.48} & \textbf{26.64} & \textbf{63.35} & \textbf{53.13} \\
1 hop &  6.23 &  5.86 & 52.56 & 39.84 & 54.03 & 42.77 \\
2 hop & 23.22 & 22.28 & 45.27 & 34.25 & 65.23 & 53.62 \\
3 hop & 23.19 & 21.91 & 34.07 & 24.19 & 55.65 & 44.05 \\
4 hop & 49.15 & 46.51 & 25.40 & 20.26 & 72.47 & 64.98 \\

\bottomrule
\end{tabular}
  \caption{Retrieval accuracy of agentic retrieval under different hop.}
  \label{retrieval_performance_hops_m3}
\end{table*}

\begin{table*}
  \centering
  \begin{tabular}{lcccccc}
\toprule
\multirow{2}{*}{\textbf{Model}} 
& \multicolumn{2}{c}{\textbf{Text}} 
& \multicolumn{2}{c}{\textbf{Image}} 
& \multicolumn{2}{c}{\textbf{All}} \\

\cmidrule(lr){2-3}
\cmidrule(lr){4-5}
\cmidrule(lr){6-7}
& \textbf{Page} & \textbf{Section}
& \textbf{Page} & \textbf{Section}
& \textbf{Page} & \textbf{Section} \\
\midrule

Heuristic Retrieval & 29.85 & 27.64 & 29.56 & 21.27 & 57.21 & 46.92 \\
Agentic Retrieval & 30.30 & 28.73 & 35.48 & 26.64 & 63.35 & 53.13 \\
\bottomrule
\end{tabular}
  \caption{Retrieval accuracy of the heuristic retrieval and agentic retrieval.}
  \label{retrieval_performance}
\end{table*}

Based on the attribution annotations in our dataset, we can assess whether retrieval-augmented models successfully retrieve the correct information. We provide more detailed retrieval accuracy results for the heuristic retrieval and agentic retrieval in Tables~\ref{retrieval_performance_hops} and \ref{retrieval_performance_hops_m3}. ``Text" refers to text-based retrieval, ``Image" refers to image-based retrieval, and ``All represents the combined results of both. ``Page" indicates the accuracy of retrieving the correct entity page, while ``Section" denotes the accuracy of retrieving the correct section containing the evidence. ``Top-k" refers to the number of hops in the retrieved evidence. ``all" in Hop refers to all hop levels.

As seen in both tables, the accuracy of question-based text retrieval for 1 hop queries is extremely low, primarily because 1 hop questions do not contain additional textual entities. The highest text retrieval accuracy is observed for 4+ hop queries, largely because higher hop questions tend to include more textual entities. On the other hand, although the number of visual entities increases with the number of hops, the accuracy of image retrieval decreases, reflecting the inherent difficulty of retrieving information involving multiple visual entities.

Comparing the results in Table~\ref{retrieval_performance}, we observe that agentic retrieval retrieves the correct knowledge base pages approximately 6\% more often than heuristic retrieval. At the more fine-grained level of knowledge base sections, its retrieval accuracy is also about 6\% higher. This indicates that decomposing complex queries into single hop queries can effectively improve retrieval precision. More specifically, we observe that agentic retrieval achieves 1\% higher accuracy in text retrieval and 5\% higher accuracy in image retrieval compared to the heuristic approach. Therefore, the segmentation of multi-entity images contributes more significantly to performance improvement than the segmentation of text queries.

Additionally, we find that agentic retrieval performs worse than heuristic retrieval in image retrieval accuracy for 1 hop queries. This is understandable, as single entity images do not require segmentation, and the segmentation of agentic retrieval introduces additional noise in these cases. The improvements in the image retrieval accuracy are more pronounced at 3 hop and 4+ hop levels compared to 2 hop, further demonstrating that entity segmentation is more effective when dealing with images containing more visual entities.

\subsection{The Impact of Evidence Number}

\begin{table*}
  \centering
  \small
  \begin{tabular}{clccccccccc}
\toprule
\multirow{2}{*}{\textbf{Setting}} & \multirow{2}{*}{\textbf{Model}} 
& \multicolumn{4}{c}{\textbf{Hop}} 
& \multicolumn{4}{c}{\textbf{Entity}} 
& \multirow{2}{*}{\textbf{All}}  \\
\cmidrule(lr){3-6}
\cmidrule(lr){7-10}
& & \textbf{1} & \textbf{2} & \textbf{3} & \textbf{4+} 
  & \textbf{1} & \textbf{2} & \textbf{3} & \textbf{4+} 
  &  \\

\midrule
\multirow{5}{*}{Gold@0}
& Qwen2.5-VL-7B-Instruct         & 34.87 & 20.66 & 21.63 & 21.74 & 
24.14 & 24.30 & 21.08 & 21.74 & 22.53  \\
& Qwen2.5-VL-72B-Instruct        & 47.94 & 29.48 & 34.19 & 29.20 & 
30.81 & 35.51 & 36.16 & 29.20 & 32.55  \\
& InternVL2.5-8B                & 26.14 & 16.86 & 19.44 & 17.36 & 
18.45 & 19.60 & 19.91 & 17.36 & 18.69  \\
& MiniCPM-V-2.6                 & 31.46 & 19.13 & 23.05 & 25.05 & 
19.17 & 23.05 & 25.31 & 25.05 & 23.45  \\
& Llama-3.1-8B-Instruct          & 26.16 & 15.94 & 21.66 & 21.54 & 
18.62 & 19.27 & 21.95 & 21.54 & 20.61  \\

\midrule
\multirow{5}{*}{Gold@1}
& Qwen2.5-VL-7B-Instruct         & 67.77 & 32.43 & 33.69 & 28.13 & 
39.01 & 38.23 & 35.73 & 28.13 & 34.40  \\
& Qwen2.5-VL-72B-Instruct        & 71.78 & 40.44 & 42.10 & 33.97 & 
43.02 & 48.04 & 45.20 & 33.97 & 41.49  \\
& InternVL2.5-8B                & 60.63 & 32.41 & 33.34 & 27.92 & 
33.54 & 39.92 & 36.61 & 27.92 & 33.61  \\
& MiniCPM-V-2.6                 & 64.30 & 33.16 & 32.60 & 31.82 & 
34.87 & 40.60 & 35.94 & 31.82 & 35.12  \\
& Llama-3.1-8B-Instruct          & 61.05 & 27.63 & 26.48 & 22.95 & 
34.35 & 32.35 & 27.88 & 22.95 & 28.48  \\

\midrule
\multirow{5}{*}{Gold@2}
& Qwen2.5-VL-7B-Instruct         & 67.77 & 54.52 & 39.18 & 34.75 & 
49.28 & 54.98 & 43.53 & 34.75 & 43.86  \\
& Qwen2.5-VL-72B-Instruct        & 71.78 & 64.18 & 51.24 & 38.64 & 
54.37 & 66.75 & 56.77 & 38.64 & 51.93  \\
& InternVL2.5-8B                & 60.63 & 52.24 & 37.93 & 33.43 & 
44.53 & 54.18 & 42.05 & 33.43 & 41.81  \\
& MiniCPM-V-2.6                 & 64.30 & 51.45 & 37.94 & 36.11 & 
44.94 & 53.67 & 42.53 & 36.11 & 42.84  \\
& Llama-3.1-8B-Instruct          & 61.05 & 46.53 & 33.28 & 26.69 & 
46.43 & 45.58 & 35.05 & 26.69 & 36.71  \\

\midrule
\multirow{5}{*}{Gold@3}
& Qwen2.5-VL-7B-Instruct         & 67.77 & 54.52 & 49.52 & 36.65 & 
60.50 & 54.98 & 47.06 & 36.65 & 47.97  \\
& Qwen2.5-VL-72B-Instruct        & 71.78 & 64.18 & 63.86 & 39.50 & 
63.57 & 66.75 & 65.10 & 39.50 & 56.63  \\
& InternVL2.5-8B                & 60.63 & 52.24 & 46.89 & 36.64 & 
51.34 & 54.18 & 47.62 & 36.64 & 45.84  \\
& MiniCPM-V-2.6                 & 64.30 & 51.45 & 46.34 & 37.44 & 
55.24 & 53.67 & 44.47 & 37.44 & 46.29  \\
& Llama-3.1-8B-Instruct          & 61.05 & 46.53 & 40.57 & 31.78 & 
53.93 & 45.58 & 38.05 & 31.78 & 40.95  \\

\midrule
\multirow{5}{*}{Gold@All}
& Qwen2.5-VL-7B-Instruct        & 67.77 & 54.52 & 49.52 & 39.52 & 
60.51 & 54.98 & 47.06 & 39.52 & 48.97  \\
& Qwen2.5-VL-72B-Instruct       & 71.78 & 64.18 & 63.86 & 45.01 & 
63.46 & 66.75 & 65.10 & 45.01 & 58.41  \\
& InternVL2.5-8B               & 60.63 & 52.24 & 46.89 & 39.26 & 
51.36 & 54.18 & 47.62 & 39.26 & 46.82  \\
& MiniCPM-V-2.6                & 64.30 & 51.45 & 46.34 & 39.77 & 
55.27 & 53.67 & 44.47 & 39.77 & 46.92  \\
& Llama-3.1-8B-Instruct         & 61.05 & 46.53 & 40.57 & 36.13 & 
53.79 & 45.58 & 38.05 & 36.13 & 42.26  \\

\bottomrule
\end{tabular}
  \caption{Performance comparison of various models under different evidence number.}
  \label{model_performance_additional_evidence}
\end{table*}

In multi-hop, multi-entity reasoning tasks, the model must rely on multiple interrelated pieces of knowledge to complete intermediate reasoning chains and gradually construct a logical path toward the final answer. To further analyze the role of evidence in such complex tasks, we designed a controlled experiment: we provided the model with varying numbers of gold evidence for each question and observed how the accuracy changed. The results for 4-hop and 4-entity cases have already been discussed in the main text. Table~\ref{model_performance_additional_evidence} presents more detailed results.

\section{Case Study}

\begin{figure*}
  \centering
  \includegraphics[width=0.9\textwidth]{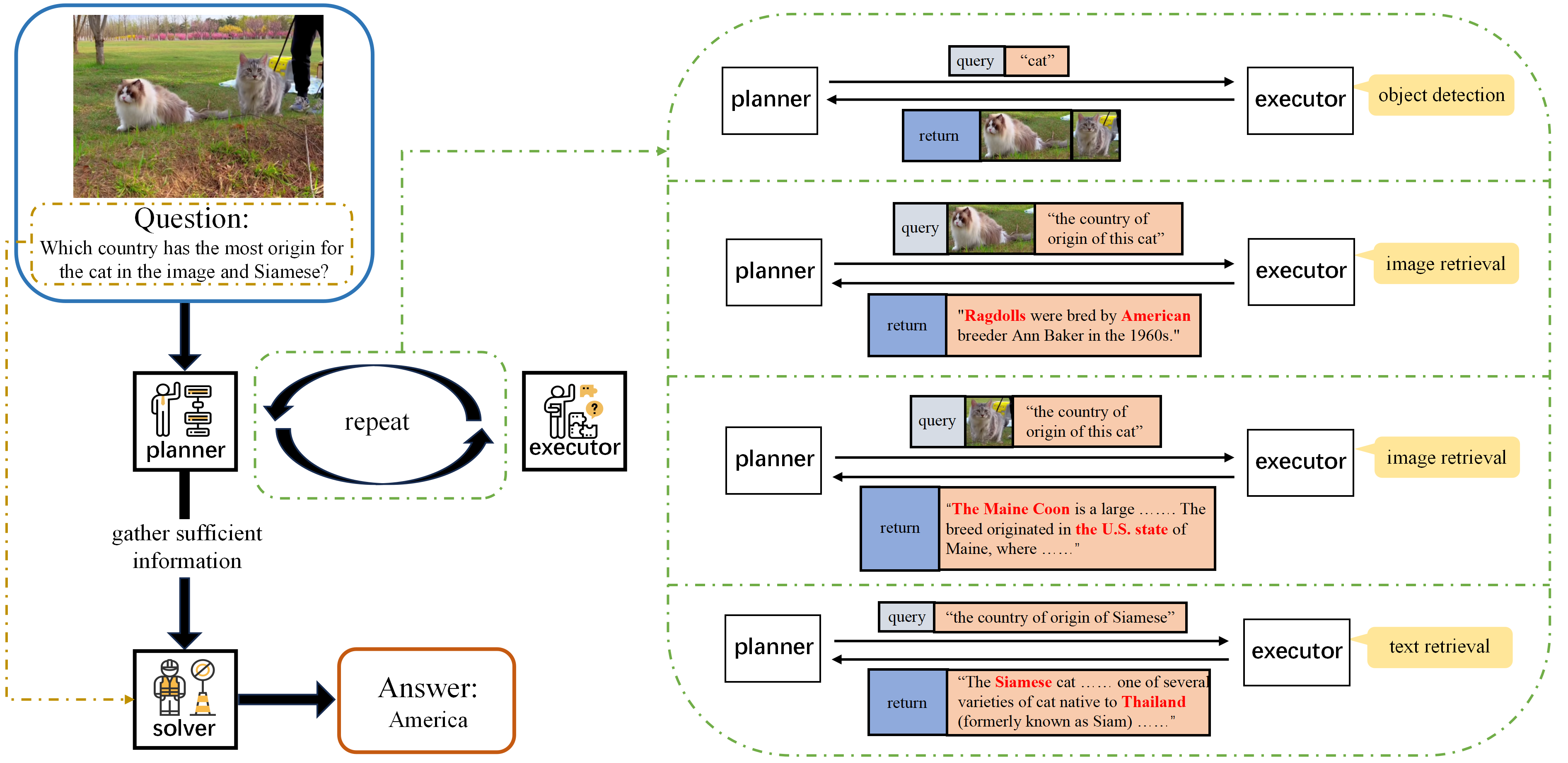}
  \caption{Schematic diagram and case study of the agentic retrieval model.}
  \label{case_study}
\end{figure*}

We provide a case study of agentic retrieval in Figure~\ref{case_study}. The input image contains two cats on the grass. The input question is: \textit{“Which country has the most origin for the cat in the image and Siamese?”} This means that we need to (1) identify the breeds of the two cats in the image, (2) retrieve the origin countries for both cats and the Siamese breed, and (3) determine which country appears most frequently among those origins. Since there are two visual entities in the image and one textual entity in the question, we have: $IP = 2, TP = 1, P = 3, S = 1, hop = 3$.

In agentic retrieval, the Planner first invokes the Object Detection module in the executor to segment the two cats from the image. Then, it calls the Image Retrieval module twice to obtain information about the origins of the two cats. Afterwards, it uses the Text Retrieval module to obtain information about the origin of the Siamese breed. Before each call to the executor, the planner must assess whether sufficient information has already been acquired. If not, it continues the loop by calling the executor again. Once the planner deems the information sufficient, it organizes the retrieved data into an evidence list and passes it to the Solver. The solver then reasons over the image, the question, and the evidence to derive the final answer: \textit{America}.

\section{Compute Resources}
Experiments were conducted on an internal computing cluster equipped with NVIDIA A800 GPUs. The total computational cost was approximately 2000 GPU hours, primarily used for dataset construction and model inference. To accelerate inference and improve throughput, we deployed the models using the vLLM \cite{kwon2023efficient} framework, which enables efficient execution of large language models while optimizing GPU memory usage. No large-scale model training was performed during the course of this study.

\section{Dataset Examples}
We provide several dataset examples for readers to review intuitively.

\begin{figure*}
\centering
\begin{tcolorbox}[enhanced, colback=white, colframe=black, title={Example 1},sharp corners, boxrule=0.5mm]

\begin{center}
    \includegraphics[width=0.6\textwidth]{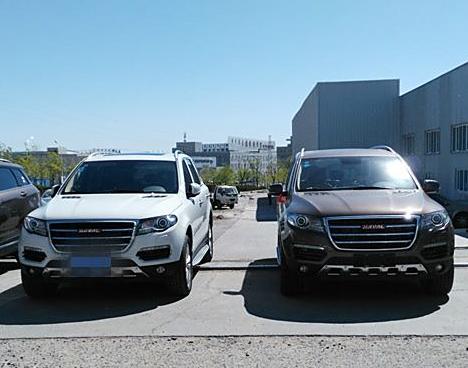}
\end{center}

\textbf{Question:} Which manufacturer is the most common among all the vehicle models shown in the image, Volvo Xc60 and Hyundai Accent?

\textbf{Question Type:} IP2TP2S1 \hspace{1em}
\textbf{Question Hop:} 4 \hspace{1em}
\textbf{Entity Num:} 4 \hspace{1em}\\
\textbf{Answers:} Great Wall. \hspace{1em} \hspace{1em} \\
\textbf{Image Entity Names:} Haval H8, Haval H8 \\
\textbf{Evidence:}
\begin{itemize}
\item The Haval H8 is a mid-size SUV produced by Haval, a sub-brand of Great Wall Motor.
\item The Volvo XC60 is a compact luxury crossover SUV manufactured and marketed by Swedish automaker Volvo Cars since 2008.
\item The Hyundai Accent, or Hyundai Verna is a subcompact car produced by Hyundai.
\end{itemize}

\textbf{Evidence URLs:}
\begin{itemize}
    \item \url{https://en.wikipedia.org/wiki/Haval_H8}
    \item \url{https://en.wikipedia.org/wiki/Volvo_XC60}
    \item \url{https://en.wikipedia.org/wiki/Hyundai_Accent}
\end{itemize}
\end{tcolorbox}
\end{figure*}

\begin{figure*}
\begin{tcolorbox}[enhanced, colback=white, colframe=black, title={Example 2},sharp corners, boxrule=0.5mm]

\begin{center}
    \includegraphics[width=0.6\textwidth]{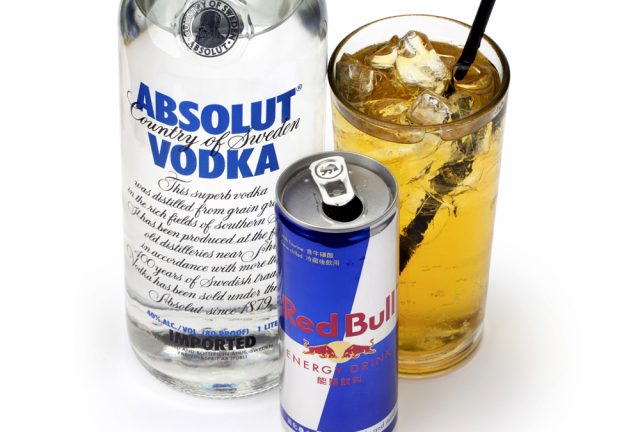}
\end{center}

\textbf{Question:} Which of the brands in the picture, Hyundai Motor Company and Doritos was established earliest?

\textbf{Question Type:} IP2TP2S1 \hspace{1em}
\textbf{Question Hop:} 4 \hspace{1em}
\textbf{Entity Num:} 4 \hspace{1em}\\
\textbf{Answers:} Absolut Vodka. \hspace{1em} \\
\textbf{Image Entity Names:} Absolut Vodka, Red Bull \\
\textbf{Evidence:}
\begin{itemize}
\item Absolut was established in 1879 by Lars Olsson Smith and is produced in Åhus, Sweden.
\item Since its launch in 1987, more than 100 billion cans of Red Bull have been sold worldwide, including 9.8 billion in 2021.
\item Hyundai Motor Company was founded in 1967.
\item Doritos is an American brand of flavored tortilla chips produced since 1964 by Frito-Lay, a wholly owned subsidiary of PepsiCo.
\end{itemize}

\textbf{Evidence URLs:}
\begin{itemize}
    \item \url{https://en.wikipedia.org/wiki/Absolut_Vodka}
    \item \url{https://en.wikipedia.org/wiki/Red_Bull}
    \item \url{https://en.wikipedia.org/wiki/Hyundai_Motor_Company}
    \item \url{https://en.wikipedia.org/wiki/Doritos}
\end{itemize}
\end{tcolorbox}
\end{figure*}

\begin{figure*}
\begin{tcolorbox}[enhanced, colback=white, colframe=black, title={Example 3},sharp corners, boxrule=0.5mm]

\begin{center}
    \includegraphics[width=0.6\textwidth]{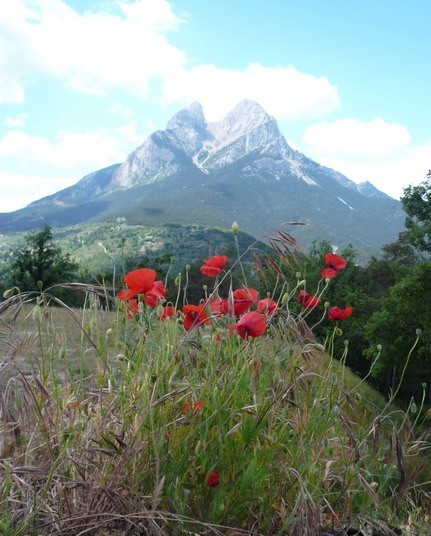}
\end{center}

\textbf{Question:} What is the conservation status of the animal that is a popular dish in the village nearest to this mountain's west?

\textbf{Question Type:} IP1TP0S3 \hspace{1em}
\textbf{Question Hop:} 3 \hspace{1em}
\textbf{Entity Num:} 1 \hspace{1em}\\
\textbf{Answers:} Least Concern. \hspace{1em} \\
\textbf{Image Entity Names:} Pedraforca \\
\textbf{Evidence:}
\begin{itemize}
\item Located within the Cadí-Moixeró Natural Park, Pedraforca has been declared a Natural Site of National Interest by the Generalitat de Catalunya. 
The closest villages to Pedraforca are Gósol to the west and Saldes to the east. Pedraforca marks the boundary between the two municipalities, as well as between the provinces of Barcelona and Lleida.
\item Pèsol negre, a local variety of 'black pea'; Blat de moro escairat, or 'peeled corn', often cooked in a pork broth; Patates emmascarades, or "Masked Potatoes", mashed potatoes cooked with blood or black pudding; All i oli with pork, eaten during the swine-harvest in fall; Veal with wild mushrooms; Wild boar
\item It has been assessed as least concern on the IUCN Red List due to its wide range, high numbers, and adaptability to a diversity of habitats.
\end{itemize}

\textbf{Evidence URLs:}
\begin{itemize}
    \item \url{https://en.wikipedia.org/wiki/Pedraforca}
    \item \url{https://en.wikipedia.org/wiki/G%C3%B3sol}
    \item \url{https://en.wikipedia.org/wiki/Wild_boar}
\end{itemize}
\end{tcolorbox}
\end{figure*}

\begin{figure*}
\begin{tcolorbox}[enhanced, colback=white, colframe=black, title={Example 4},sharp corners, boxrule=0.5mm]

\begin{center}
    \includegraphics[width=0.6\textwidth]{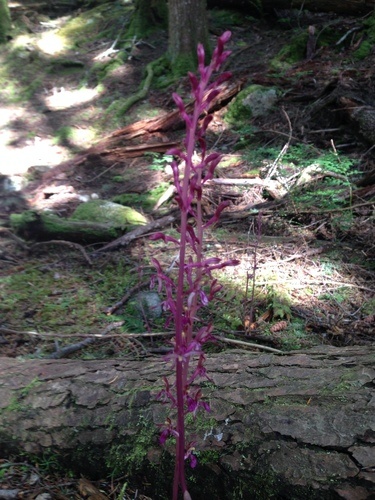}
\end{center}

\textbf{Question:} Where is the highest point of the mountains in which this plant is found?

\textbf{Question Type:} IP1TP0S2 \hspace{1em}
\textbf{Question Hop:} 2 \hspace{1em}
\textbf{Entity Num:} 1 \hspace{1em}\\
\textbf{Answers:} Mount Rainier. \hspace{1em} \\
\textbf{Image Entity Names:} Corallorhiza mertensiana \\
\textbf{Evidence:}
\begin{itemize}
\item Corallorrhiza mertensiana is found in the Cascades from Alaska to California, and the Rocky Mountains from Alberta to Wyoming.
\item The highest peak in the range is Mount Rainier in Washington at 14,411 feet (4,392 m).
\end{itemize}

\textbf{Evidence URLs:}
\begin{itemize}
    \item \url{https://en.wikipedia.org/wiki/Corallorhiza_mertensiana}
    \item \url{https://en.wikipedia.org/wiki/Cascade_Range}
\end{itemize}
\end{tcolorbox}
\end{figure*}

\begin{figure*}
\begin{tcolorbox}[enhanced, colback=white, colframe=black, title={Example 5},sharp corners, boxrule=0.5mm]

\begin{center}
    \includegraphics[width=0.6\textwidth]{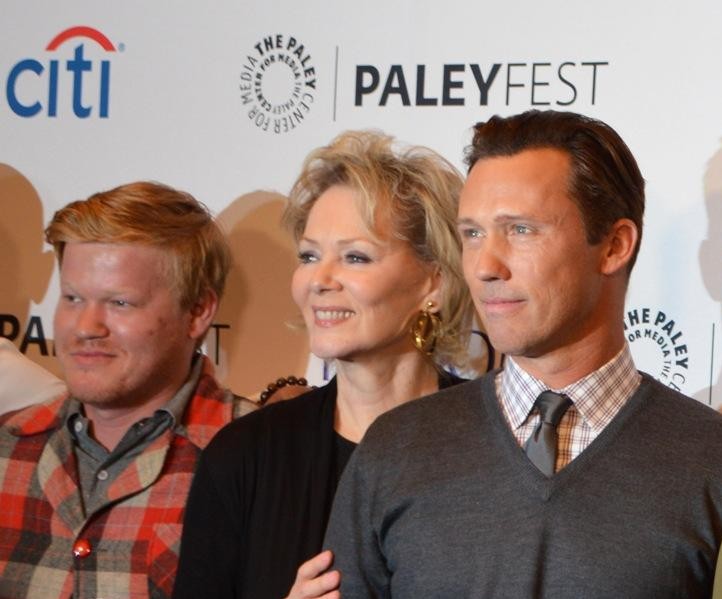}
\end{center}

\textbf{Question:} Did all the people in the photo come from the same country of birth?

\textbf{Question Type:} IP3TP0S1 \hspace{1em}
\textbf{Question Hop:} 3 \hspace{1em}
\textbf{Entity Num:} 3 \hspace{1em}\\
\textbf{Answers:} Yes. \\
\textbf{Image Entity Names:} Jesse Plemons, Jean Smart, Jeffrey Donovan \\
\textbf{Evidence:}
\begin{itemize}
\item Jesse Plemons is an American actor.
\item Jean Elizabeth Smart (born September 13, 1951) is an American actress.
\item Jeffrey Donovan (born May 11, 1968) is an American actor.
\end{itemize}

\textbf{Evidence URLs:}
\begin{itemize}
    \item \url{https://en.wikipedia.org/wiki/Jesse_Plemons}
    \item \url{https://en.wikipedia.org/wiki/Jean_Smart}
    \item \url{https://en.wikipedia.org/wiki/Jeffrey_Donovan}
\end{itemize}
\end{tcolorbox}
\end{figure*}

\begin{figure*}
\begin{tcolorbox}[enhanced, colback=white, colframe=black, title={Example 6},sharp corners, boxrule=0.5mm]

\begin{center}
    \includegraphics[width=0.6\textwidth]{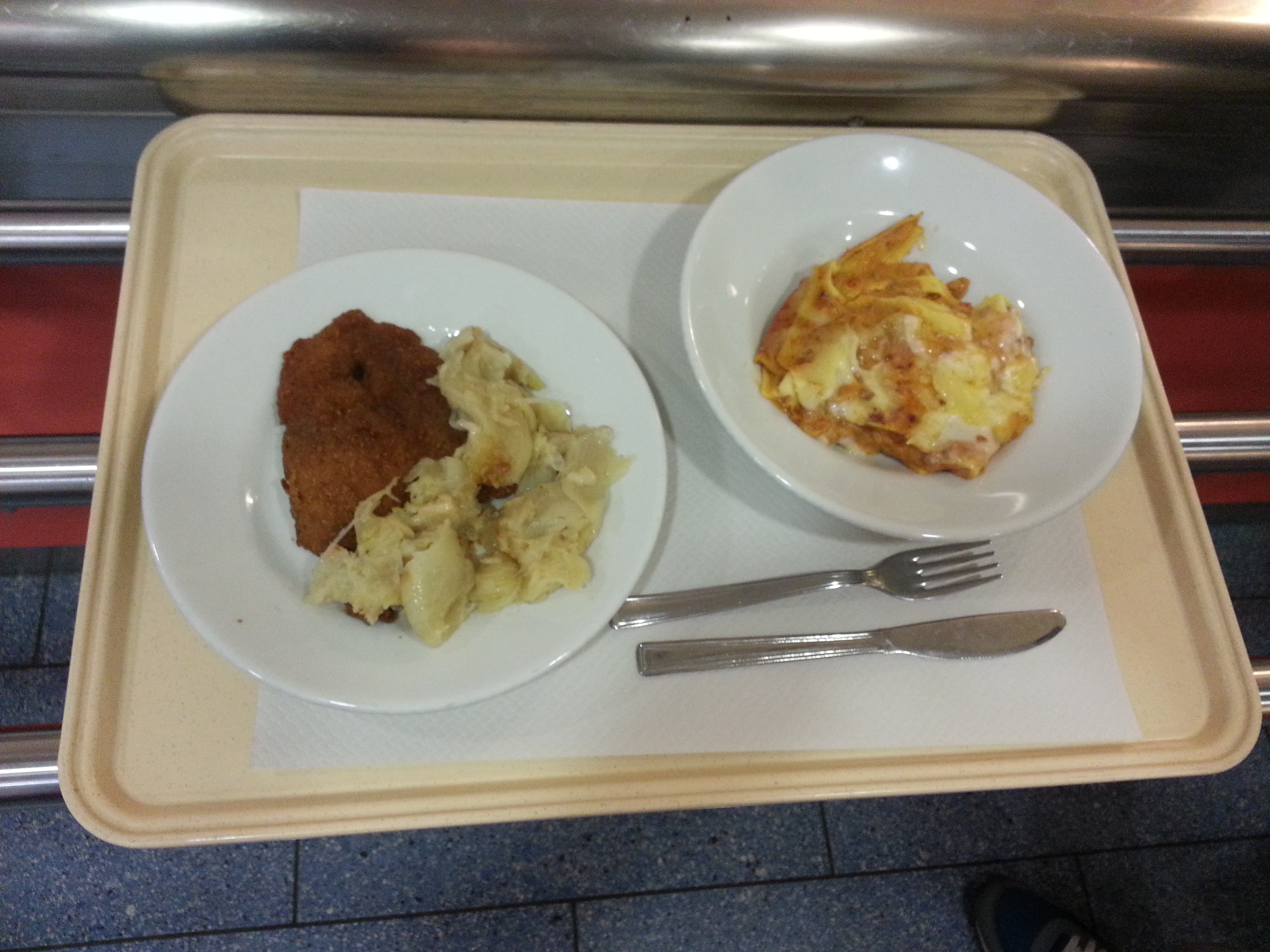}
\end{center}

\textbf{Question:} Which country is most associated with the origin of the food in the image?

\textbf{Question Type:} IP2TP0S1 \hspace{1em}
\textbf{Question Hop:} 2 \hspace{1em}
\textbf{Entity Num:} 2 \hspace{1em}\\
\textbf{Answers:} Italy. \\
\textbf{Image Entity Names:} Cotoletta, lasagna \\
\textbf{Evidence:}
\begin{itemize}
\item Cotoletta alla Bolognese is a traditional dish of Bologna.
\item Lasagne are a type of pasta, possibly one of the oldest types, made of very wide, flat sheets.
\end{itemize}

\textbf{Evidence URLs:}
\begin{itemize}
    \item \url{https://en.wikipedia.org/wiki/Cotoletta}
    \item \url{https://en.wikipedia.org/wiki/Lasagne}
\end{itemize}
\end{tcolorbox}
\end{figure*}

\begin{figure*}
\begin{tcolorbox}[enhanced, colback=white, colframe=black, title={Example 7}, sharp corners, boxrule=0.5mm]

\begin{center}
    \includegraphics[width=0.6\textwidth]{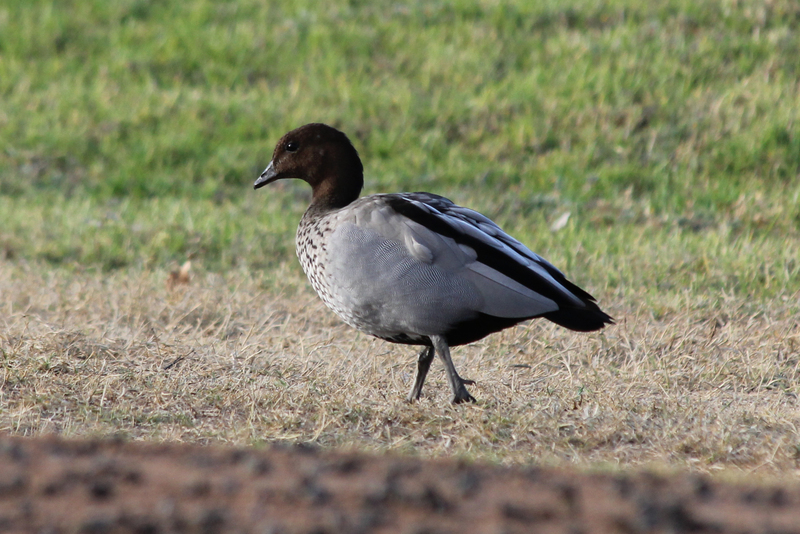}
\end{center}

\textbf{Question:} Where are this bird and Paradise shelduck native to? \\
\textbf{Question Type:} IP1TP1S1 \hspace{1em}
\textbf{Question Hop:} 2 \hspace{1em}
\textbf{Entity Num:} 2 \hspace{1em}\\
\textbf{Answers:} Australia, New Zealand. \\
\textbf{Image Entity Names:} Australian wood duck \\
\textbf{Evidence:}
\begin{itemize}
    \item The Australian wood duck, maned duck or maned goose (\textit{Chenonetta jubata}) is a dabbling duck found throughout much of Australia.
    \item The paradise shelduck (\textit{Tadorna variegata}), also known as the paradise duck, or pūtangitangi in Māori, is a species of shelduck, a group of goose-like ducks, which is endemic to New Zealand.
\end{itemize}

\textbf{Evidence URLs:}
\begin{itemize}
    \item \url{https://en.wikipedia.org/wiki/Australian_wood_duck}
    \item \url{https://en.wikipedia.org/wiki/Paradise_shelduck}
\end{itemize}
\end{tcolorbox}
\end{figure*}

\begin{figure*}
\begin{tcolorbox}[enhanced, colback=white, colframe=black, title={Example 8},sharp corners, boxrule=0.5mm]

\begin{center}
    \includegraphics[width=0.6\textwidth]{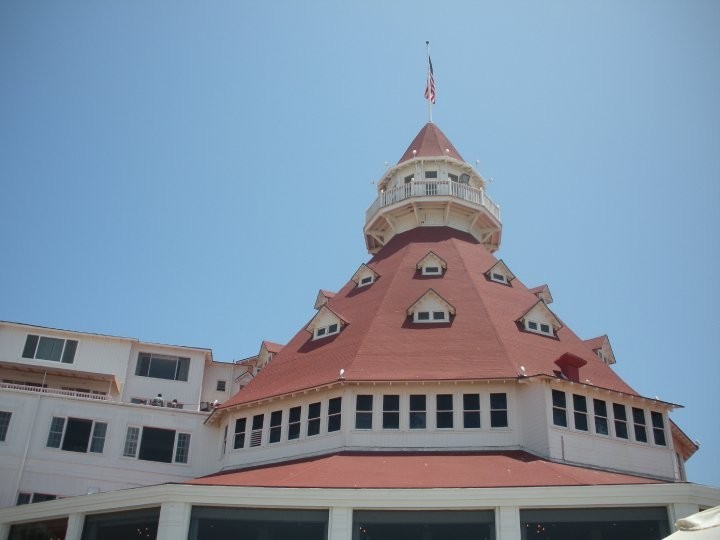}
\end{center}

\noindent
\textbf{Question:} Who are the owners of this building, Providence Park and Oxburgh Hall?\\
\textbf{Question Type:} IP1TP2S1 \hspace{1em}
\textbf{Question Hop:} 3 \hspace{1em}
\textbf{Entity Num:} 3 \hspace{1em}\\
\textbf{Answers:} Anbang Insurance Group, Portland, National Trust.\\
\textbf{Image Entity Names:} Hotel del Coronado \\
\textbf{Evidence:}
\begin{itemize}
    \item In March 2016, Blackstone sold Strategic Hotels \& Resorts to Anbang Insurance Group, a Beijing-based Chinese insurance company, in a \$6.5 billion deal involving multiple resorts.
    \item Providence Park (formerly Jeld-Wen Field; PGE Park; Civic Stadium; originally Multnomah Stadium; and from 1893 until the stadium was built, Multnomah Field) is an outdoor soccer venue located in the Goose Hollow neighborhood of Portland, Oregon.
    \item The Bedingfelds gained the manor of Oxborough through marriage in the early 15th century, and the family has lived at the hall since its construction, although ownership passed to the National Trust in 1952.
\end{itemize}

\textbf{Evidence URLs:}
\begin{itemize}
    \item \url{https://en.wikipedia.org/wiki/Hotel_del_Coronado}
    \item \url{https://en.wikipedia.org/wiki/Providence_Park}
    \item \url{https://en.wikipedia.org/wiki/Oxburgh_Hall}
\end{itemize}

\end{tcolorbox}
\end{figure*}

\end{document}